\documentclass[preprint,12pt,3p]{elsarticle}

\usepackage{amssymb}

\usepackage{booktabs}
\usepackage{multirow}
\usepackage{subcaption}
\usepackage{algorithm}
\usepackage{algorithmic}
\usepackage{threeparttable}
\usepackage{hyperref}

\def\ie{{\em i.e.}}
\def\eg{{\em e.g.}}

\journal{Nuclear Physics B}

\begin{document}

\begin{frontmatter}

\title{Towards Better Object Detection in Scale Variation with \\ Adaptive Feature Selection}

\author[label1]{Zehui Gong}
\ead{zehuigong@foxmail.com}

\author[label1]{Dong Li\corref{cor1}}  
\address[label1]{Guangdong University of Technology}
\ead{dong.li@gdut.edu.cn}
\cortext[cor1]{Dong Li is the corresponding author.}


\begin{abstract}
It is a common practice to exploit pyramidal feature representation to tackle the problem of scale variation in object instances. However, most of them still predict the objects in a certain range of scales based solely or mainly on a single-level representation, yielding inferior detection performance. To this end, we propose a novel adaptive feature selection module (AFSM), to automatically learn the way to fuse multi-level representations in the channel dimension, in a data-driven manner. It significantly improves the performance of the detectors that have a feature pyramid structure, while introducing nearly free inference overhead. Moreover, we propose a class-aware sampling mechanism to tackle the class imbalance problem, which automatically assigns the sampling weight to each of the images during training, according to the number of objects in each class. Experimental results demonstrate the effectiveness of the proposed method, with 83.04\% mAP at 15.96 FPS on the VOC dataset, and 39.48\% AP on the VisDrone-DET validation subset, respectively, outperforming other state-of-the-art detectors considerably. The code is available at \url{https://github.com/ZeHuiGong/AFSM.git}.

\end{abstract}

\begin{keyword}
object detection \sep scale variation \sep multi-scale fusion
\end{keyword}

\end{frontmatter}


\section{Introduction}
\label{intro}
\begin{figure}[tbp]
	\centering
	\begin{minipage}[t]{0.48\linewidth}
		\centering
		\includegraphics[width=0.95\linewidth, height=0.6\linewidth]{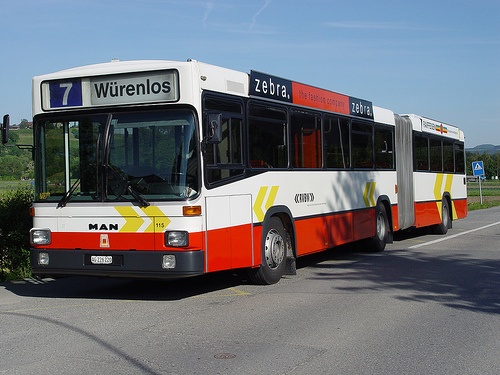}
		\subcaption{large objects}
		\label{large_object}
	\end{minipage}
	\begin{minipage}[t]{0.48\linewidth}
		\centering
		\includegraphics[width=0.95\linewidth, height=0.6\linewidth]{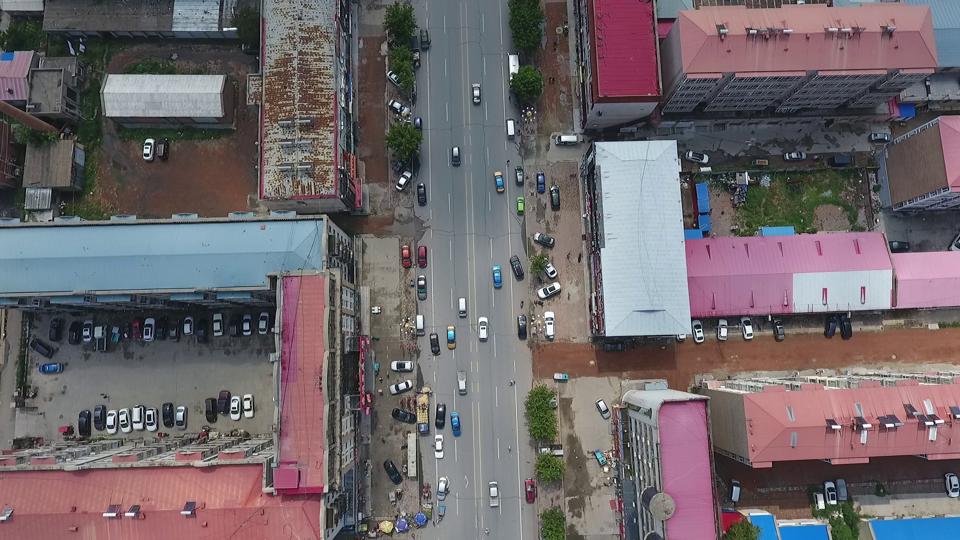}
		\subcaption{small objects}
		\label{small_object}
	\end{minipage}
	\caption{The size of objects varies significantly, posing a great challenge for object detection algorithms. In (a), the large object almost occupies 80\% of an image, while in (b), the small ones can hardly be clearly captured without zooming.}
	\label{scale_variation}
\end{figure}

In recent years, the performance of object detection consecutively improves, thanks to the carefully designed deep convolutional neural networks (CNNs) \cite{he2016deep, huang2017densely, simonyan2014very, szegedy2015going} and high-quality large-scale benchmarks \cite{lin2014microsoft, pascal_voc}. However, these advanced detectors \cite{ren2015faster, redmon2016yolo, lin2017focal} are still struggling to tackle the wide range of variation in object scale, in real-world applications (see Fig. \ref{scale_variation}). To solve this problem, recent researches focus mainly on two possible strategies: image pyramid \cite{singh2018analysis} and feature pyramid \cite{liu2016ssd, lin2017feature, he2017mask}. 

As for the image pyramid, a set of resized versions for one input image are sequentially sent to the detector at the testing phase. Therefore, the object scale will fall into an appropriate range at a specific image pyramid. However, this will increase the computational cost considerably, making it unsuitable for real-world applications, especially for mobile devices. Conversely, another line of research constructs the feature pyramid based on the inherent multi-scale feature maps on the backbone network, which is memory efficient and consumes less computation. SSD \cite{liu2016ssd} directly uses the multi-scale feature maps generated by the bottom-up pathway. FPN \cite{lin2017feature} constructs the feature pyramid by sequentially fusing high-level semantic information into low-level feature maps, with an additional top-down pathway and lateral connection. 

Although FPN and other similar works \cite{fu2017dssd, woo2018stairnet, zhang2018single, redmon2018yolov3} achieve satisfactory performance, they still suffer from the following limitation. Specifically, each object in a specific range of scale is predicted based solely or mainly on the feature from a single level, while discarding the information in other levels, resulting in losing the chance to capture richer information. Generally, High-level features contain richer context and semantic information, with large receptive fields. This may be useful for small object prediction because the context information can be an additional cue for inferring the class of the small objects. Similarly, low-level features are spatially finer and, contain more details for high localization accuracy. Allowing large objects to access them is undoubtedly beneficial. It is hence achieving inferior detection performance to predict object instance in a certain range of size, based solely on single-level features. Some models attempt to solve this problem by enabling multi-level feature fusion. ASFF \cite{liu2019learning} fuses multi-level features via spatially point-wise addition. SFAM \cite{zhao2019m2det} re-weights the concatenated features at each scale with a SE module \cite{hu2018squeeze}.

Another challenge for robust object detection is the problem of class imbalance, \ie, the number of objects varies significantly from class to class, resulting in two sets of classes, the major and minor classes. This posts great challenges for the detection algorithms, since the training process can be dominated by the major classes, leading to rarely supervisory signals for the minor ones and consequently, harming the overall performance. Therefore, it is crucial to devise an effective mechanism to tackle the class imbalance problem.

In this paper, we propose a novel and effective adaptive feature selection module (AFSM), to tackle the problem of significant variation in object scale. The proposed method learns to adaptively select useful information from multi-level representations in the channel dimension. Specifically, for the feature maps at a certain level, we first resize the feature maps from other levels into the same resolution. Then, the multi-level features are integrated by the corresponding selection weight (a learnable vector). Therefore, at each channel dimension, features at each level have a corresponding importance weight, thus automatically enhancing the discriminative clues beneficial to the following detection. The proposed approach is agnostic to the detector, \ie, irrespective of one-stage or two-stage, and it can be applied to the detectors that have multi-level representations. Moreover, we propose a class-aware sampling mechanism (CASM) to solve the problem of class imbalance, which automatically assigns the sampling weight to each of the training images, according to the number of objects in each class. CASM increase the occurrence frequency of the minor class objects, resulting in more supervisory signals for these objects, hence improving the performance of minor classes considerably, as shown in Section \ref{sec_abla}.

The main contributions of this paper are summarized as follows:
\begin{itemize}
	\item[1.] We propose an adaptive feature selection module to adaptively select useful information across multi-scale representations, which enhances the ability of the model to cope with the variation in object scales.
	\item[2.] Noticing that the problem of class imbalance restricts the performance of the detector, we propose a class-aware sampling mechanism to tackle the class imbalance problem, which can automatically assign the sampling weight to each of the images during training, according to the number of objects in each class.
	\item[3.] Comprehensive experiments conducted on the VisDrone and PASCAL-VOC datasets confirm the effectiveness of our proposed method, which achieves state-of-the-art accuracy compared to other detectors, while still running in real-time.
\end{itemize}

In Section \ref{relat}, we review the related works of object detection and feature pyramid structure. The details of the proposed algorithm are illustrated in Section \ref{method}. And we describe the implementation details and the experimental results in Section \ref{exps}. Finally, a brief conclusion of our work is given in Section \ref{conclusion}.

\section{Related Work}
\label{relat}
\noindent\textbf{Object detection} Formally, the modern object detection methods can be categorized into two types: one-stage pipeline \cite{redmon2016yolo, liu2016ssd, lin2017focal} and two-stage pipeline \cite{cai2018cascade, ren2015faster, wu2019double-head}. One-stage detectors achieve real-time performance by outputting the detection results directly without proposal generation procedure. Specifically, YOLO \cite{redmon2016yolo} regresses bounding boxes and class probabilities in a single efficient backbone network. SSD \cite{liu2016ssd} covers objects of various sizes by adjusting the output bounding boxes on the default boxes over multiple feature maps. RetinaNet \cite{lin2017focal} settles the extreme foreground-background class imbalance, thus achieving better performance than two-stage object detectors. Alternatively, two-stage detectors obtain higher accuracy by combining the region proposal stage and classification stage. Specifically, Faster R-CNN \cite{ren2015faster} introduces a region proposal network (RPN), which simultaneously predicts bounding boxes and objectness scores to achieved further speeds-up. Double-Head RCNN \cite{wu2019double-head} handles the classification in a fully connected head while regressing the bounding in a convolution head. Recently, some researches \cite{law2018cornernet, duan2019centernet, zhou2019objects, Zhou2019BottomUpOD} provide a different view of objects, by detecting keypoints of the bounding box. 

\noindent\textbf{Feature pyramid} Feature pyramid representations, \ie, multi-level feature maps, is a basic solution to tackle the variation in object scale. Skip connection \cite{long2015fully} and hyper-column \cite{hariharan2015hypercolumns} are the early attempts to integrate multi-level feature to generate more representative feature maps. After that, multi-scale prediction \cite{liu2016ssd, lin2017feature, liu2018path, tan2020efficientdet} begins to emerge, because of the lightweight property, higher accuracy, and faster inference compared to image pyramids. SSD \cite{liu2016ssd} predicts class scores and bounding boxes respectively at multiple feature levels. FPN \cite{lin2017feature} introduces a top-down pathway, and a lateral connection to sequentially merge high-level semantic information into the low-level features, building multi-level representations for the following detection head. PANet \cite{liu2018path} adds an extra bottom-up pathway on the basis of FPN to shorten the distance between high-level and low-level features. In BiFPN \cite{tan2020efficientdet}, multiple PANet-like structures with scale-wise level re-weighting is proposed, to fuse feature maps from different scales. Another line of works \cite{chen2020mnasfpn, ghiasi2019fpn} employs the Neural Architecture Search (NAS) algorithm to automatically find the powerful architecture for multi-level representations, improving the performance considerably.

\noindent\textbf{Selection of multi-scale features} Besides the generation of multi-level features, some of the researches \cite{liu2019learning, zhao2019m2det, onoro2018learning} make a step forward and, delve into the mechanism of how to select useful information across various feature levels. In \cite{onoro2018learning}, a learnable gating unit is proposed to control the flow of information in single-level skip connection, in the U-Net architecture for visual counting. Zhao et al. \cite{zhao2019m2det} propose a scale-wise feature aggregation module (SFAM) to execute channel-wise attention on the concatenated multi-scale feature maps, where the attention weights are generated by the  squeeze and excitation (SE) module \cite{hu2018squeeze}. In \cite{liu2019learning}, the point-wise attention weights are used for multi-level feature fusion, where a $1\times 1$ conv is employed to generate the weight at each level, following by a softmax layer for normalizing. 

The spatially point-wise addition proposed in \cite{liu2019learning} is not effective enough for multi-level feature selection and, introduces more computational cost (\ie, several $1\times 1$ convs for generating fusion weights). In this paper, by regarding the selection weight at each level as a learnable vector for channel-wise multi-level feature fusion, AFSM can automatically learn the distribution of object size over the whole dataset (as showed in Section \ref{sec_abla}), which makes the selection process more stable and effective. It seems that SFAM \cite{zhao2019m2det} shares the same spirit of our work. However, we emphasize that SFAM only executes channel-wise attention at a single level, on the concatenated feature maps, instead of adaptively selecting useful information across various levels.

\section{The Proposed Method}
\label{method}

\begin{figure}
	\centering
	\includegraphics[width=0.95\linewidth]{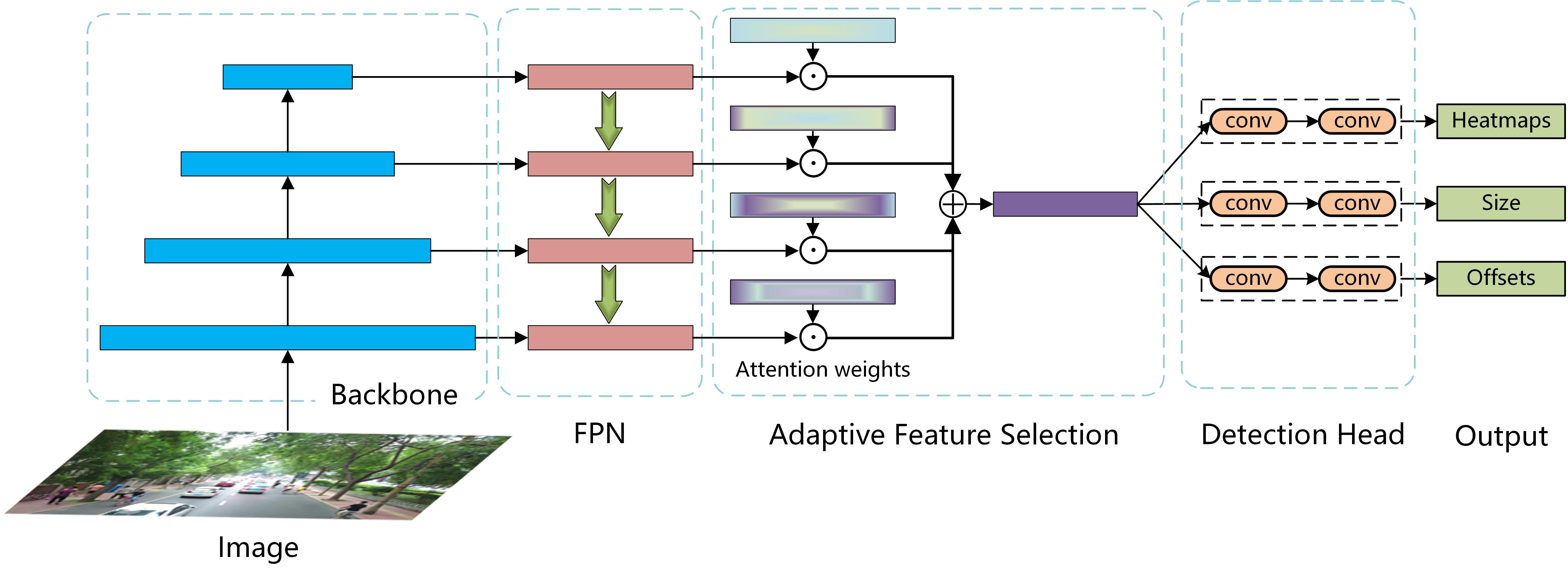}
	\caption{The overall framework of our proposed method. The backbone network is followed by the feature pyramid network (FPN) \cite{lin2017feature} for multi-scale feature generation. Then, the adaptive feature selection module is employed to adaptively select the most useful information for the subsequent detection task. Finally, the model outputs the center locations and the corresponding offsets, as well as the object sizes, to form the final bounding boxes. $\odot$ is the broadcast element-wise multiplication, and $\oplus$ means the element-wise addition. For simplicity, we show the procedure of multi-scale feature fusion in a  certain level only.}
	\label{architecture}
\end{figure}

The overall framework of our approach is shown in Fig. \ref{architecture}. In this section, we first introduce the baseline network architecture of our proposed method, which is based on the effective anchor-free detector, \ie, CenterNet \cite{zhou2019objects}. Then, we move on to push the original baseline to a stronger version, by employing the recently advanced tricks, such as mixup \cite{zhang2017mixup} and global context block (GCB) \cite{cao2019gcnet}, and by proposing two effective techniques, class aware sampling mechanism and modified GIOU \cite{rezatofighi2019generalized} loss. Finally, we describe our proposed adaptive feature selection module in detail, which adaptively selects features from multi-scale representations in the channel dimension.

\subsection{Network Architecture}
We adopt CenterNet \cite{zhou2019objects} as our baseline method, because of its simplicity and high efficiency. CenterNet presents a new representation for detecting objects, in terms of the center location of a bounding box. Other object properties, \eg, object size, and center offset, are regressed directly using the image features from the center locations. Let $(ct_x, ct_y, w, h, \epsilon_x, \epsilon_y)$ be one output that generates from the model. Then, we construct the bounding box $\mathcal{B} \in R^4$ by:

\begin{equation}
\mathcal{B}=\left \{
\begin{array}{c}
x_1 = ct_x + \epsilon_x - w / 2, \\
y_1 = ct_y + \epsilon_y - h / 2, \\
x_2 = ct_x + \epsilon_x + w / 2, \\
y_2 = ct_y + \epsilon_y + h / 2.
\end{array}
\right.
\end{equation} 

where $(ct_x, ct_y)$ is the center location of an object and, $(\epsilon_x, \epsilon_y)$ is the offset prediction used to slightly adjust the center point. $(w, h)$ is the object size prediction and, $(x_1, y_1)$ and $(x_2, y_2)$ are the top-left and bottom-right corners of the bounding box. The overall training objective of CenterNet is given as follows:

\begin{equation}
L = L_{ct} + \lambda_{size}L_{size} + \lambda_{off}L_{off}
\end{equation}

where $L_{ct}$ is the loss for center location estimation, which is a variant of the focal loss \cite{lin2017focal}. $L_{off}$ is the L1 loss for center offset regression. $L_{size}$ is a modified version of the GIOU loss that fits in the object size regression of CenterNet, and we will describe it in detail in section \ref{strong_baseline}. $\lambda_{size}$ and $\lambda_{off}$ are hyper-parameters, which are all set to 1.0 in our experiments, unless otherwise specified.

\begin{figure}[tbp]
	\centering
	\begin{minipage}[t]{\linewidth}
		\centering
		\includegraphics[width=0.48\linewidth]{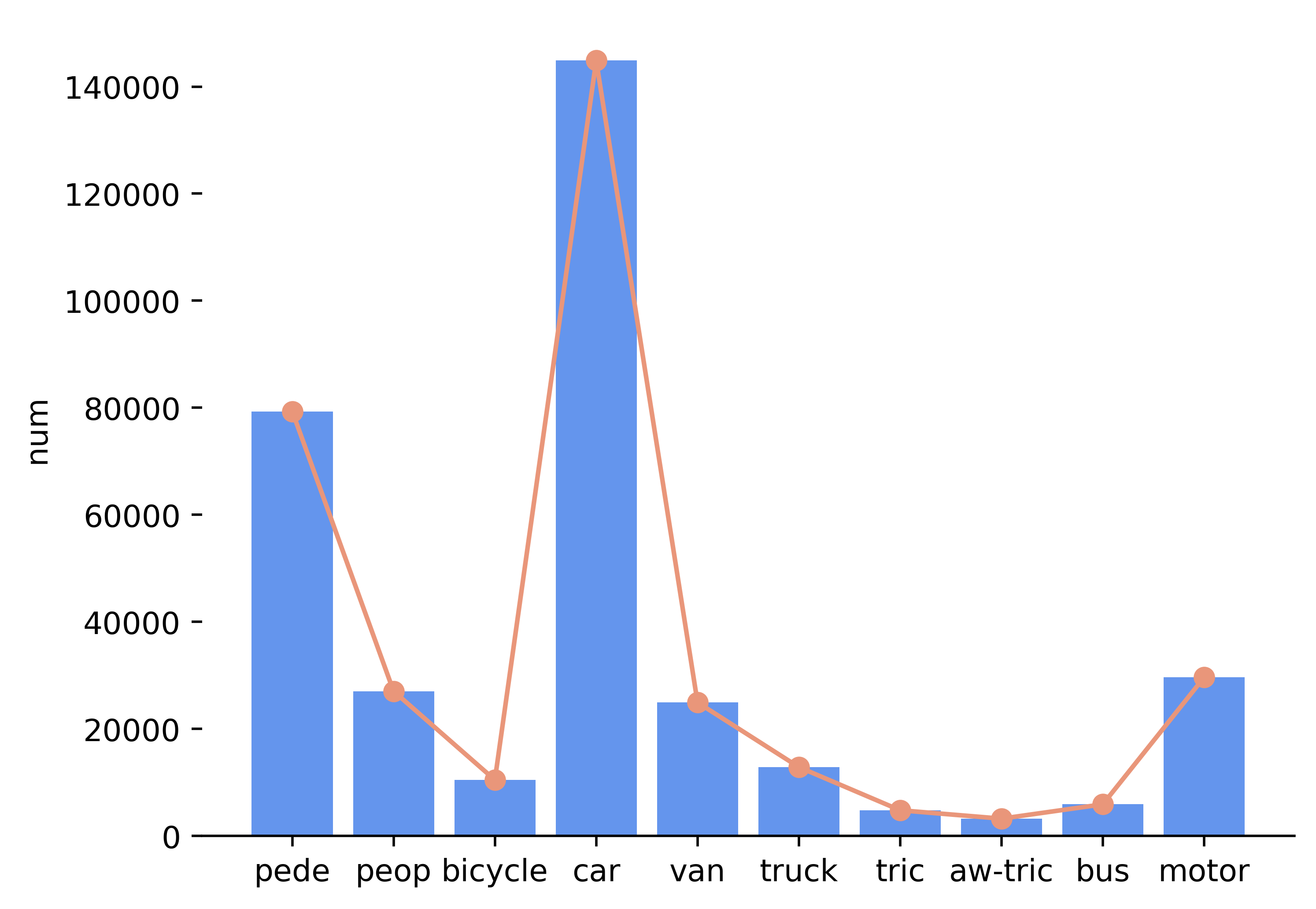}\vspace{3pt}
		\includegraphics[width=0.48\linewidth]{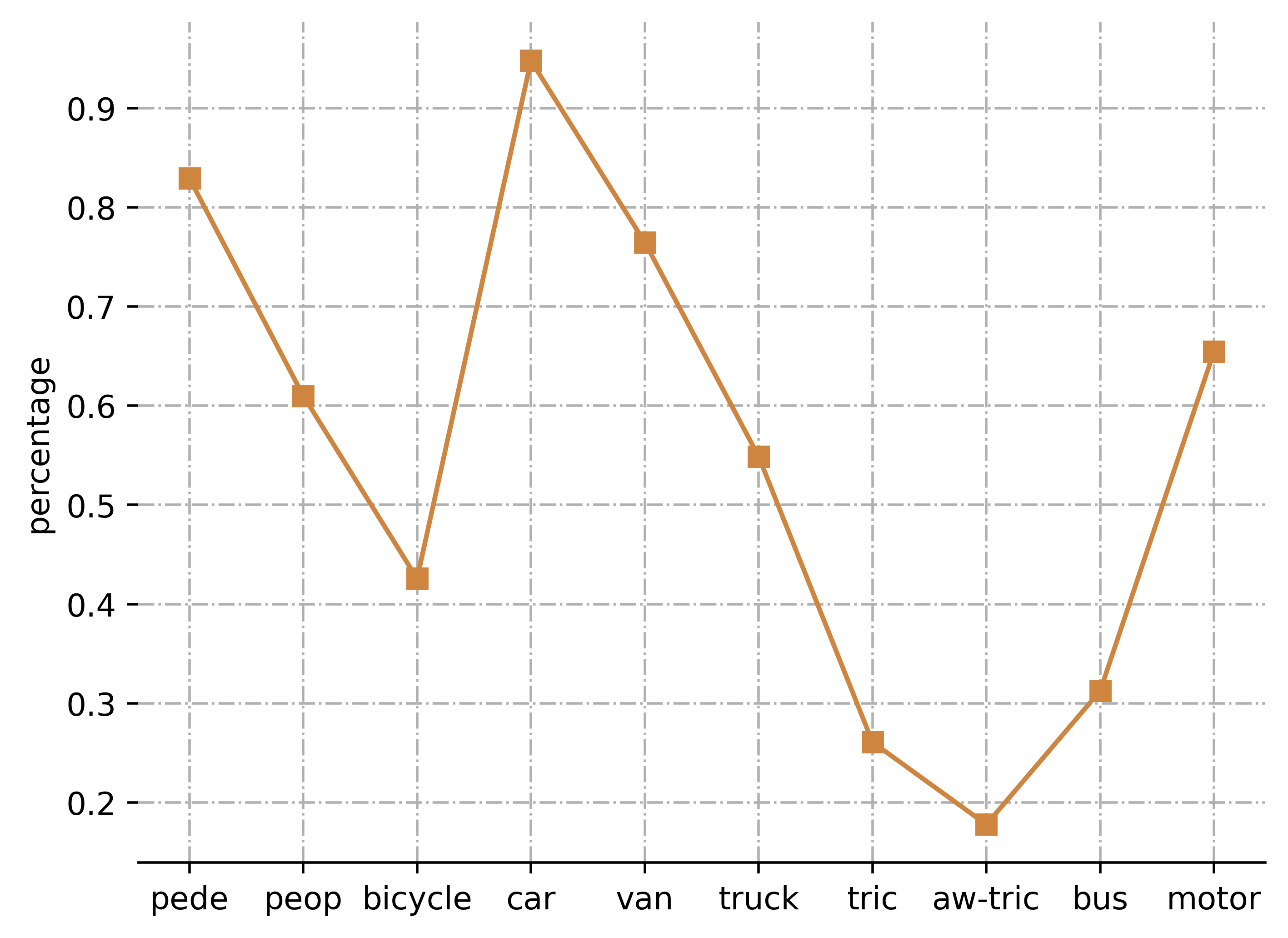}\vspace{3pt}
	\end{minipage}
	\caption{The analysis of category statistics on the VisDrone-DET training set. \textbf{Left:} The total number of object instances for each category varies significantly from class to class. \textbf{Right:} The ratio of images contains the corresponding class. The ratio of images containing `awning-tricycle' is less than 20\%}
	\label{class_statistics}
\end{figure}

\subsection{Stronger Baseline}
\label{strong_baseline}
CenterNet mainly contains two components: a backbone network (usually the Hourglass-104 \cite{newell2016stacked}) and a detection head with three parallel branches to predict the center location, offset, and object size (\ie, $w$ and $h$) of a bounding box. In this paper, we employ ResNet \cite{he2016deep} as the backbone network and, use feature pyramid network (FPN) to generate multi-scale feature maps. Moreover, we employ the recently advanced training tricks \cite{zhang2017mixup, cao2019gcnet}, and propose two effective approaches, to further improve the performance of the baseline method.

\noindent\textbf{Advanced techniques} In \cite{zhang2019bag}, mixup algorithm \cite{zhang2017mixup} is demonstrated to be an effective data augmentation method, to improve the robustness of the model by a stochastic linear interpolation between two training examples. Let $I$ be one input image, and $B=\mathrm{\{} x_1^i, y_1^i, x_2^i, y_2^i, c^i \mathrm{\}}_{i=1}^n$ denote the corresponding bounding box annotation of that image, and $n$ is the number of the boxes. Then, the mixup operation can be formulated as follows:

\begin{equation}
	\lambda_1 \sim \mathrm{Beta}(\eta, \eta),
\end{equation}
\begin{equation}
\lambda_2= \mathrm{min}(\mathrm{max}(0, \lambda_1), 1),
\end{equation}
\begin{equation}
	I=\lambda_2 I_1 + (1-\lambda_2) I_2,
\end{equation}
\begin{equation}
	B = \mathrm{concact}(B_1, B_2).
\end{equation}

\begin{algorithm}[t]
	\footnotesize
	\caption{Detailed procedures of class-aware sampling mechanism}
	\label{CASM}
	\begin{algorithmic}[1]
		\REQUIRE ~~\\ 
		The list of class labels for each of the training images, $G=\{G_i\}_{i=1}^N$, where $G_i=\{ c_j \}_{j=1}^n$, $N$ is the number of training images, $n$ the number of objects in each image. \\
		The number of training classes, $K$ \\
		\ENSURE ~~\\ 
		The list of sampling ratios for each image, $S=\{ s_1, ..., s_N \}$, used for training sampling.
		\STATE $S \leftarrow \emptyset$, $M \leftarrow \emptyset$
		\FOR{each $k \in [1, K]$}
		\STATE $M(k) \leftarrow 0$\ ~$//$ Initialize the number of objects in each class to 0.
		\ENDFOR
		\FOR{$G_i \in G$}
		\STATE $//$ Counting the number of object instance $t$ in class $k$, in image $i$
		\FOR{$k$, $t\in$ counter($G_i$)} 
		\STATE $M(k) \leftarrow M(k) + t$
		\ENDFOR
		\ENDFOR
		\FOR{each $i \in [1, N]$ }
		\STATE $s \leftarrow 0$
		\STATE $//$ Enumerate the set of unique classes in image $i$
		\FOR{$k \in \mathrm{unique}(G_i)$} 
		\STATE $s \leftarrow s + \frac{1}{M(k)}$
		\ENDFOR
		\STATE $S \leftarrow S~\cup~ \{s\}$
		\ENDFOR
		\STATE $//$ Normalize the sampling ratios
		\STATE $S \leftarrow S/\mathrm{sum}(S)$
		\RETURN $S$
	\end{algorithmic}
\end{algorithm}

Where $\mathrm{concact}(\cdot)$ means the concatenation operation of the bounding box annotation between two mixup images.

Besides, the long-range pixels in the image may be still correlated. For example, an object is less likely to be a car when the background is the sea. Therefore, it is important to model the long-range relations between pixels. We adopt the global context block (GCB) \cite{cao2019gcnet}, which is an effective module to model the global context and, is much lightweight compared to NLNet \cite{wang2018non}. 

\noindent\textbf{Class-aware training sampling} In real-world applications, imbalanced data is ubiquitous, where the number of an object instance in each class varies significantly (see Fig. \ref{class_statistics}). This posts a great challenge for the detector, since the training process can be dominated by the major classes, leading to rarely supervisory signals for the minor ones. To address this issue, we propose a novel re-weighting mechanism, \ie, class-aware sampling mechanism (CASM), to automatically assigns the sampling ratio for each image in the training set, instead of the uniform sampling. Alg. \ref{CASM} shows the detailed procedures of CASM, where the sampling ratio for each of the images is computed automatically, according to the occurrence frequency of each class in the dataset.

\noindent\textbf{Modified GIOU} \label{M_GIOU}
CenterNet uses L1 loss for object size regression, which is sensitive to the variation in object size and, separately regresses the width and height of an object, hence overlooking the correlation between properties of the same object. Conversely, GIOU loss \cite{rezatofighi2019generalized} considers the optimization as a whole, and is irrelevant to the object size. However, GIOU loss is first proposed for the anchor-based methods, which can not be applied to the CenterNet directly. Therefore, we make some modifications to the GIOU loss. The entire workflow of computing the modified GIOU (M-GIOU) loss is illustrated in Alg. \ref{MGIOU}, where an additional mask $F \in \{0,1\}^N$ is introduced to filter out the background object during back-propagation.

\begin{algorithm}[tp]
	\caption{Workflow of the modified GIOU loss}
	\label{MGIOU}
	\begin{algorithmic}[1]
		\REQUIRE
		$O^p=\{r_i^p, ..., r_N^p\}$, $O^g=\{r_i^g,...,r_N^g\}$, and $F \in \{0,1\}^N$ \\
		$O^p=\{r_i^p, ..., r_N^p\}$ is the set of predicted object size \\ $O^g=\{r_i^g,...,r_N^g\}$ is the set of ground-truth object size \\ $r_i^p=(w_i^p, h_i^p)$, $r_i^g=(w_i^g, h_i^g)$ \\
		$N$ is the maximum number of training objects in one image, \eg, $N=128$ \\
		$F$ is a mask that each element indicates the object is foreground or background.
		
		\ENSURE	$L_{size}$
		\STATE initialize $L_{size} \leftarrow 0$.
		\FOR{each $i \in [1, N]$}
		\STATE $w_i^p \leftarrow \mathrm{exp}(w_i^p)$, $h_i^p \leftarrow \mathrm{exp}(h_i^p)$. 
		\STATE Getting intersection box size:
		$t_w\leftarrow \mathrm{min}(w_i^p, w_i^g)$, $t_h\leftarrow \mathrm{min}(h_i^p, h_i^g)$.
		\STATE Getting the smallest enclosing box size: $c_w\leftarrow \mathrm{max}(w_i^p, w_i^g)$, $c_h\leftarrow \mathrm{max}(h_i^p, h_i^g)$.
		\STATE Calcalating area of $r_i^p$: $A^p\leftarrow w_i^p \times h_i^p$.
		\STATE Calculating area of $r_i^g$: $A^g\leftarrow w_i^g \times h_i^g$.
		\STATE Calculating area of smallest enclosing box: $A^c\leftarrow c_w\times c_h$.
		\STATE Calculating area of intersection box: $A^t \leftarrow t_w\times t_h$.
		\STATE $IoU \leftarrow \frac{A^t}{\mathcal{U}}$, where $\mathcal{U}=(A^p + A^g - A^t)$.
		\STATE $GIoU \leftarrow IoU - \frac{A^c - \mathcal{U}}{A^c}$.
		\STATE $//$ Multiply by the mask $F(i)$, where the loss for background object is 0.
		\STATE $L_{size} \leftarrow L_{size} + (1 - GIoU) \times F(i)$
		\ENDFOR
		\RETURN $L_{size}$.
	\end{algorithmic}
\end{algorithm}

\subsection{Adaptive Feature Selection Module}
The object scales vary significantly in real-world scenarios, posing great challenges for detection algorithms. It is hence vital to develop an approach, to cope with a wide range of variation in object scales. Therefore, we propose an adaptive feature selection module (AFSM) to adaptively learn the importance weights, for fusing feature maps from different feature levels. As shown in Fig. \ref{architecture}, the pipeline is consists of two steps: feature resizing and adaptive feature selection.

\textbf{Feature resizing.} Formally, a list of multi-scale features is given as $X=\{X^1, ..., X^l,...\}$, where $X^l$ is the feature maps at level $l$, whose resolution is $1/2^{l}$ of the input image. For feature maps at level $l$, we first resize it with a scale factor of $2^{|l-k|}$, to the same resolution of the feature maps at level $k$, where $k$ is the target level, \eg, $k=3$. As the output stride is of $4$ in CenterNet, we resize the feature maps after feature selection, to ensure the resolution is of $1/4$ of the input image. Specifically, the nearest interpolation operation is adopted to resize the features respectively at each scale.

\textbf{Adaptive feature selection.} Let $\hat{X}=\{\hat{X}^1,..., \hat{X}^l,...\}$ be the feature maps after resizing. We propose to fuse features across various resolutions as follows:

\begin{equation}
Y_i = \sum_{l} \hat{X}_i^l \cdot  \alpha_i^l,	
\end{equation}

where $Y_i$ implies the $i$-th channel of the output feature maps among spatial positions. $\alpha_i^l$ refers to the channel importance weight that controls the contribution of feature maps at level $l$ to the prediction of the bounding box. Note that $\alpha_i^l$ is a simple scalar variable, sharing across all the spatial positions. Softmax operation is employed to ensure $\sum_{l} \alpha_i^l = 1$, as follows:

\begin{figure}[tbp]
	\centering
	\begin{minipage}[t]{0.1\linewidth}
		w/o blur\vspace{1.53\linewidth}
		w/~blur\vspace{1.55\linewidth}
		mixup
	\end{minipage}
	\begin{minipage}[t]{0.28\linewidth}
		\centering
		\includegraphics[width=0.98\linewidth,height=0.63\linewidth]{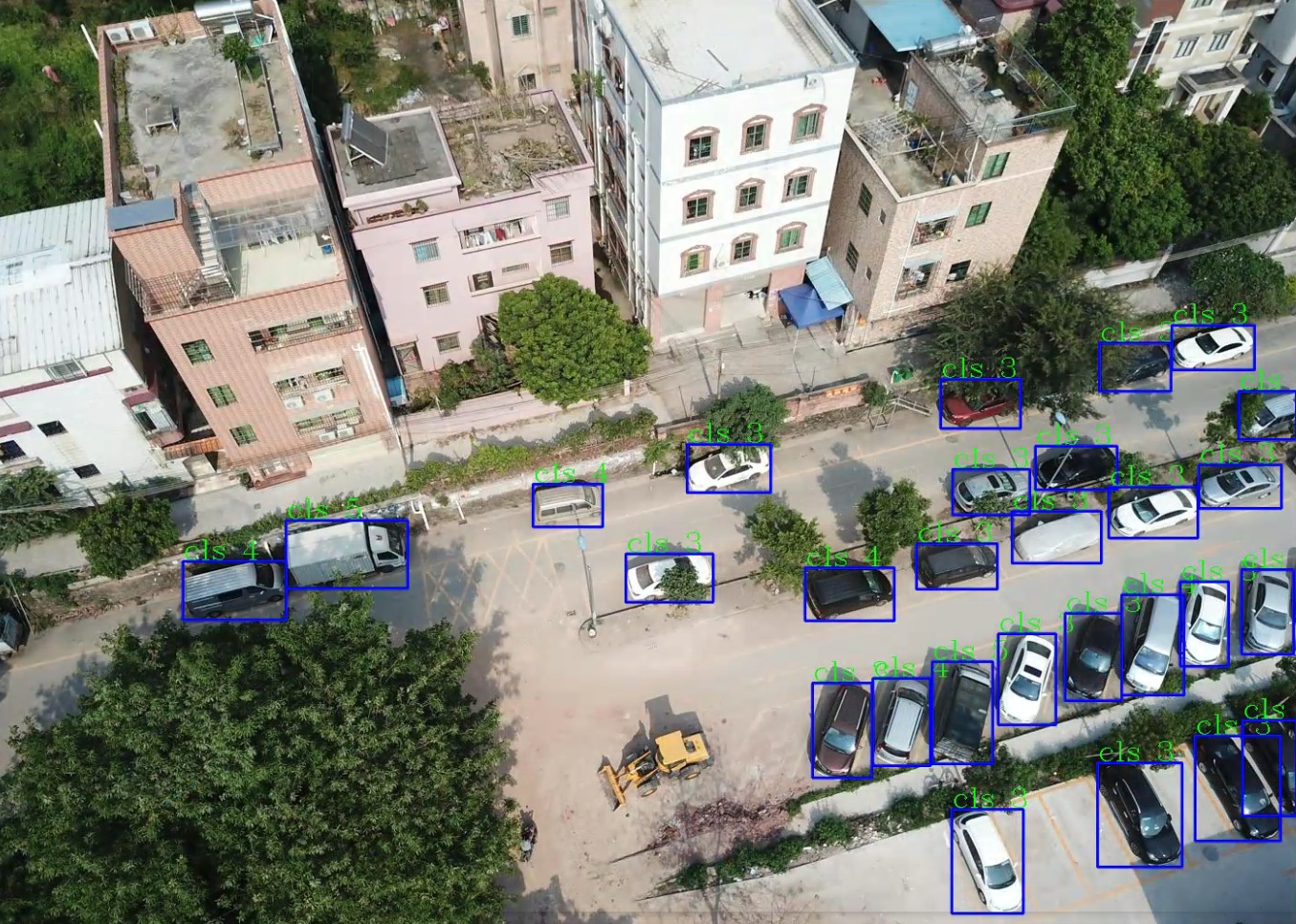}\vspace{3pt}
		\includegraphics[width=0.98\linewidth,height=0.63\linewidth]{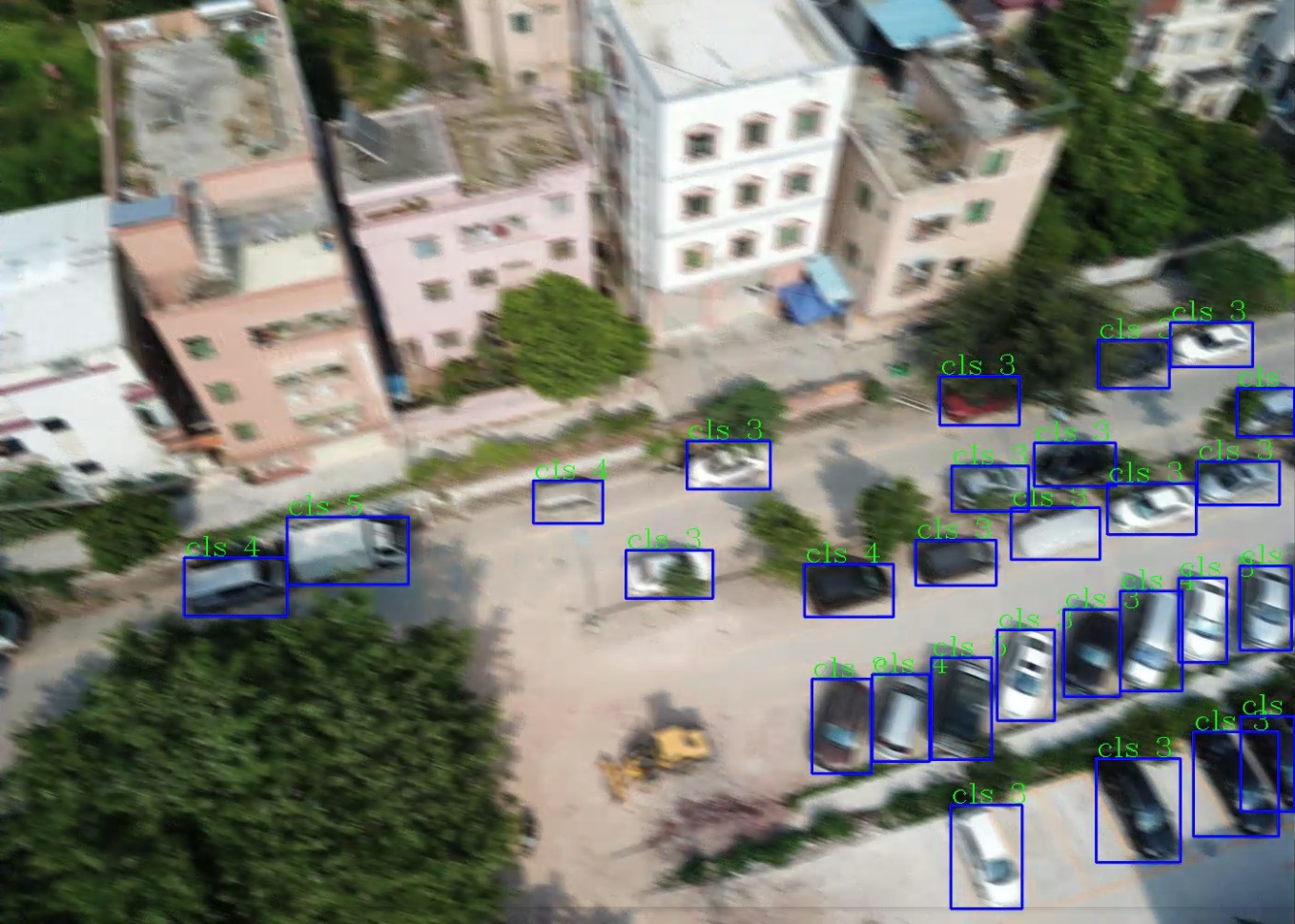}\vspace{3pt}
		
		\includegraphics[width=0.98\linewidth, height=0.63\linewidth]{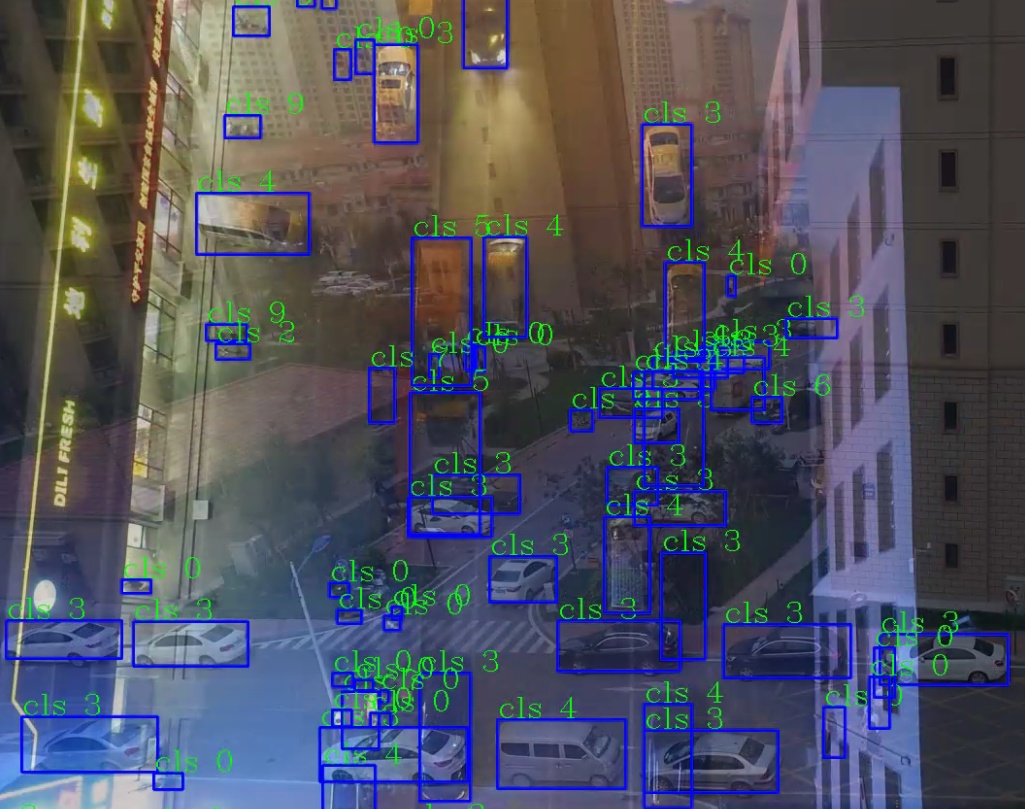}\vspace{3pt}
		
	\end{minipage}
	\begin{minipage}[t]{0.28\linewidth}
		\centering
		\includegraphics[width=0.98\linewidth,height=0.63\linewidth]{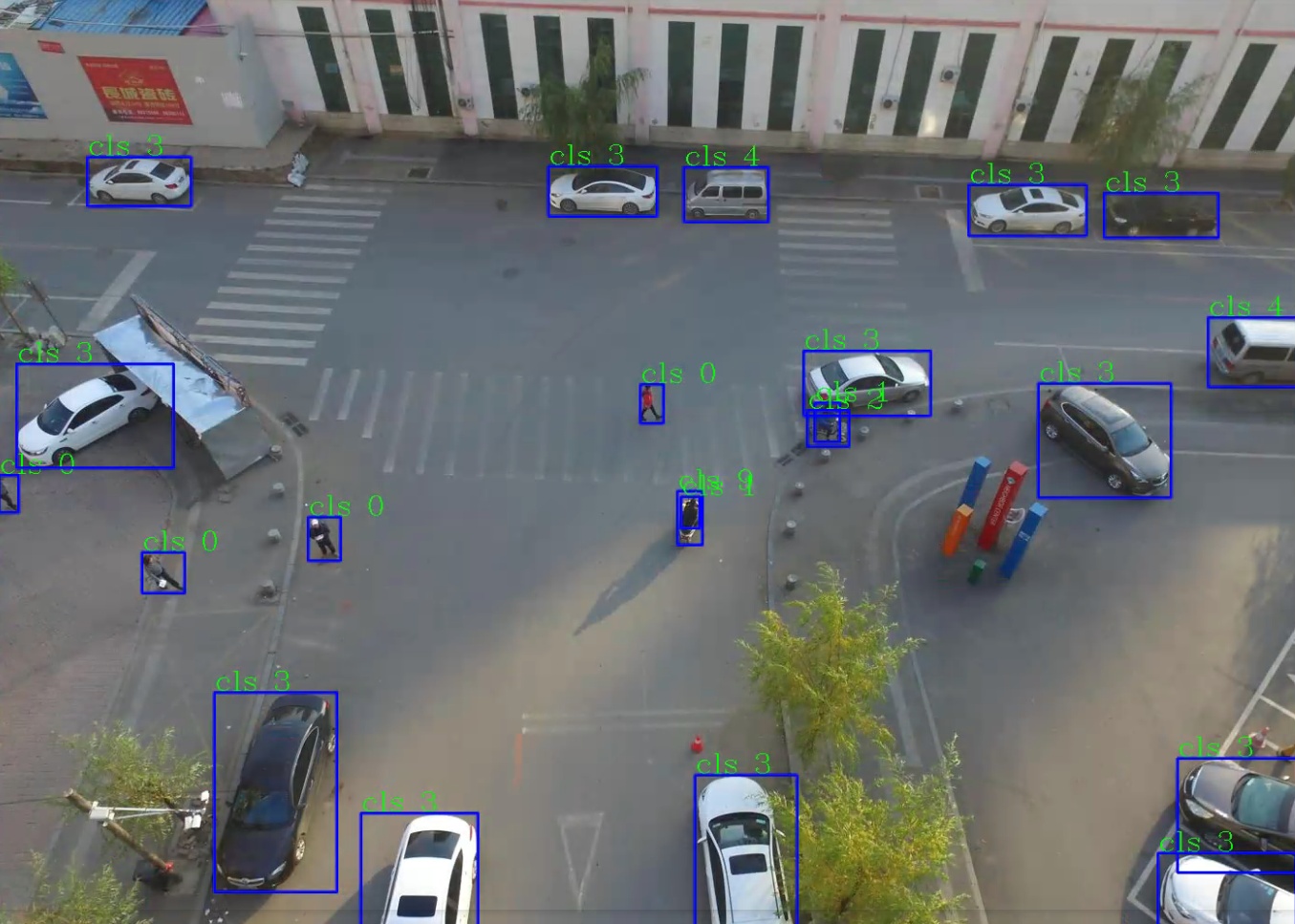}\vspace{3pt}
		\includegraphics[width=0.98\linewidth,height=0.63\linewidth]{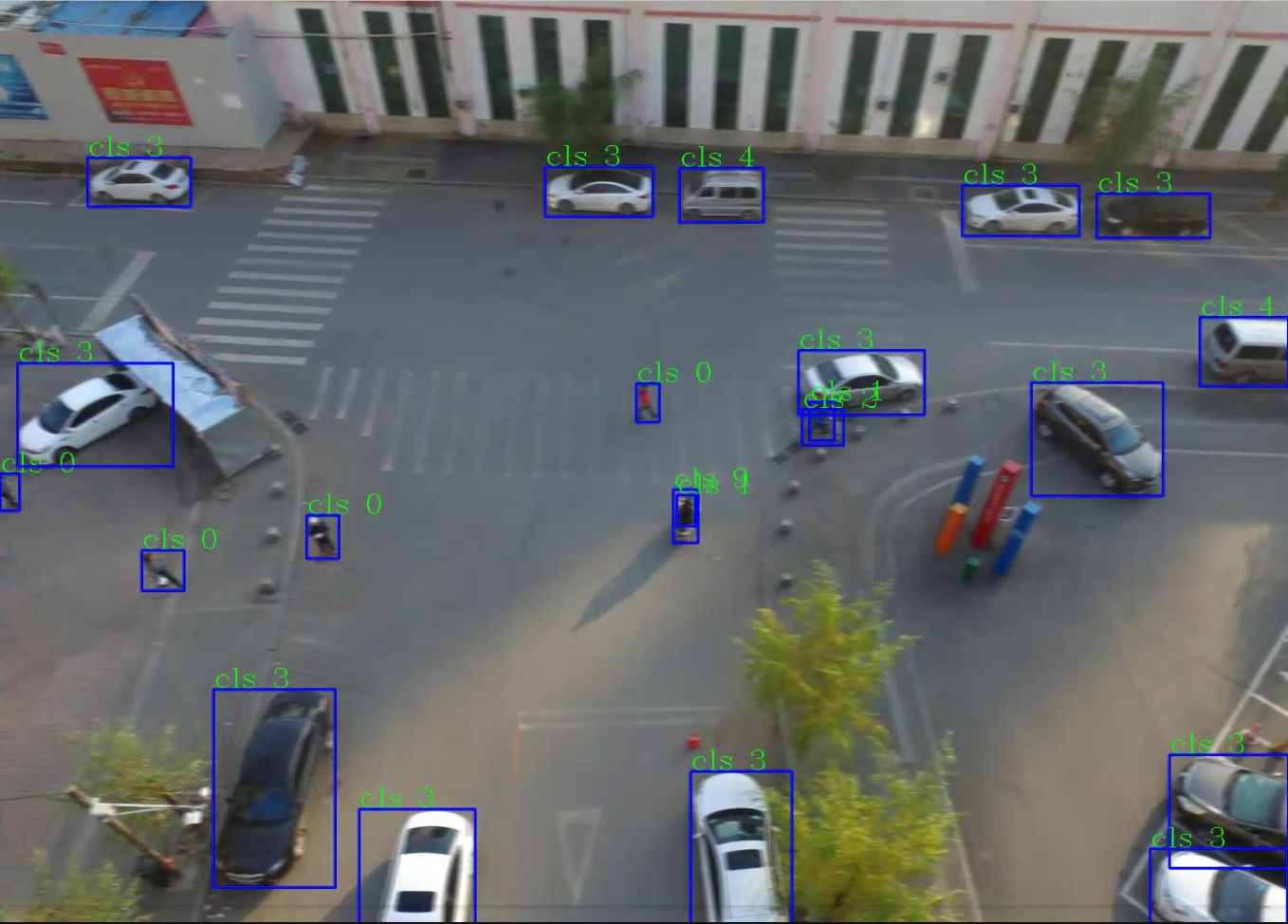}\vspace{3pt}
		\includegraphics[width=0.98\linewidth,height=0.63\linewidth]{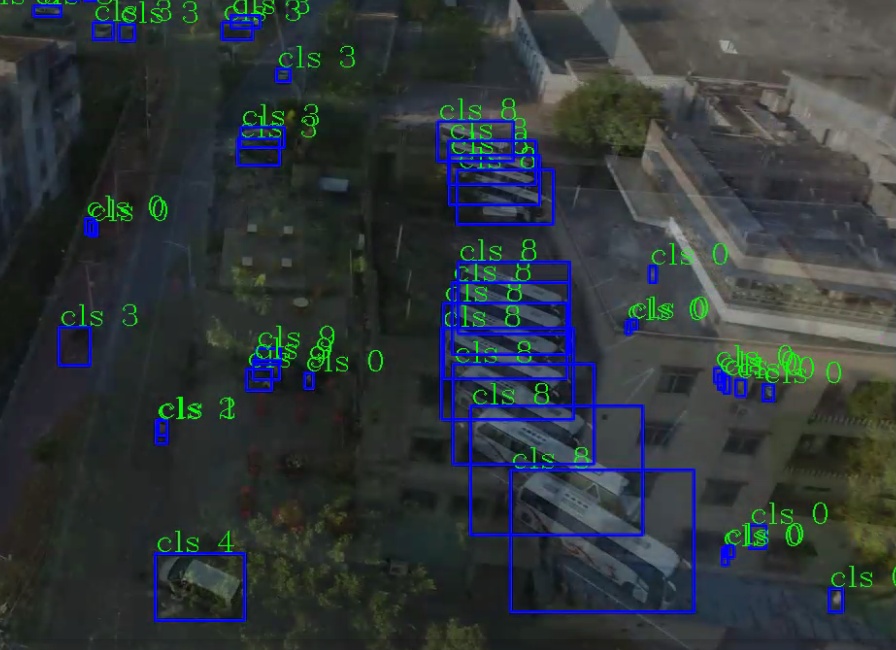}\vspace{3pt}
		
	\end{minipage}
	\begin{minipage}[t]{0.28\linewidth}
		\centering
		\includegraphics[width=0.98\linewidth,height=0.63\linewidth]{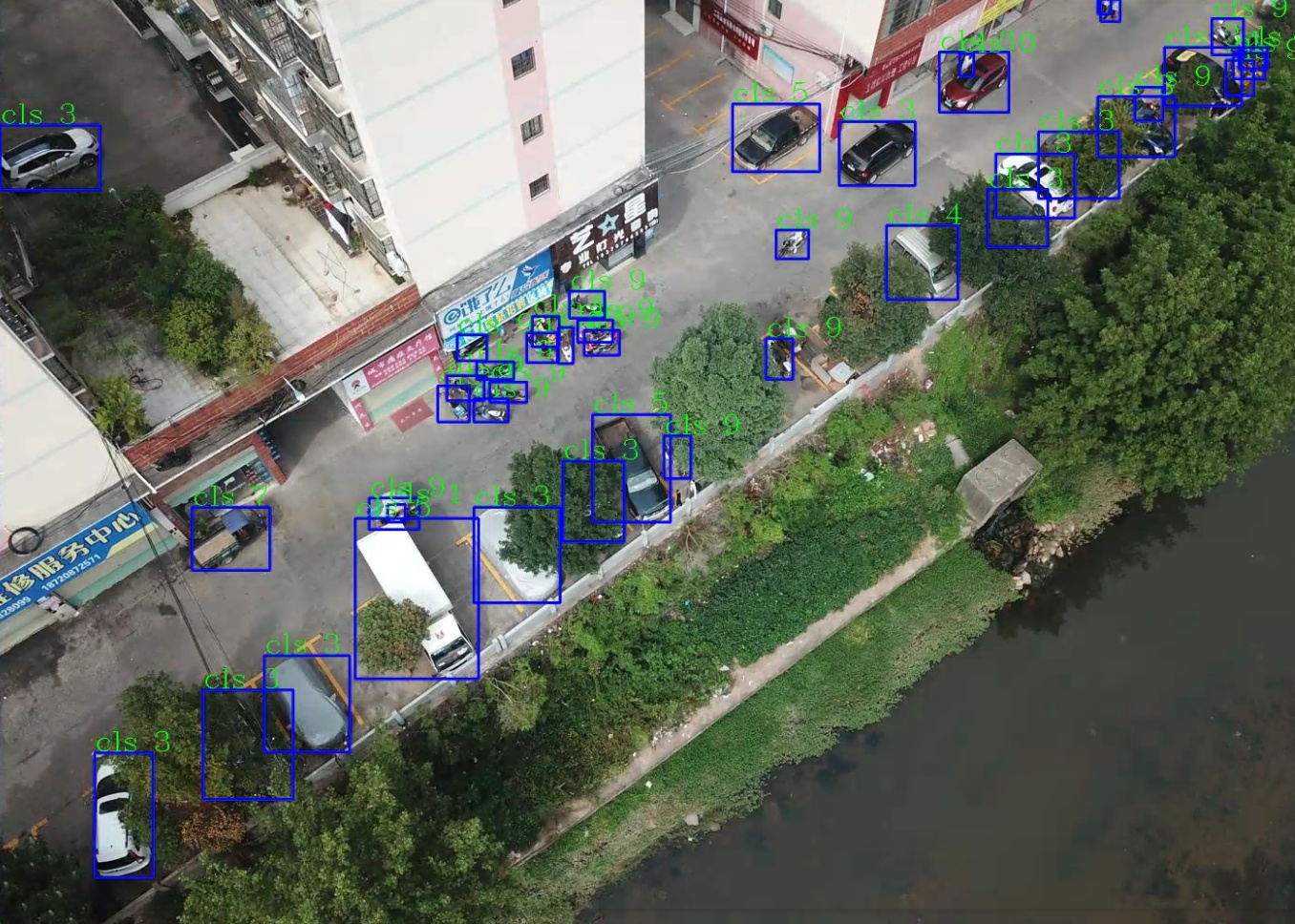}\vspace{3pt}
		\includegraphics[width=0.98\linewidth,height=0.63\linewidth]{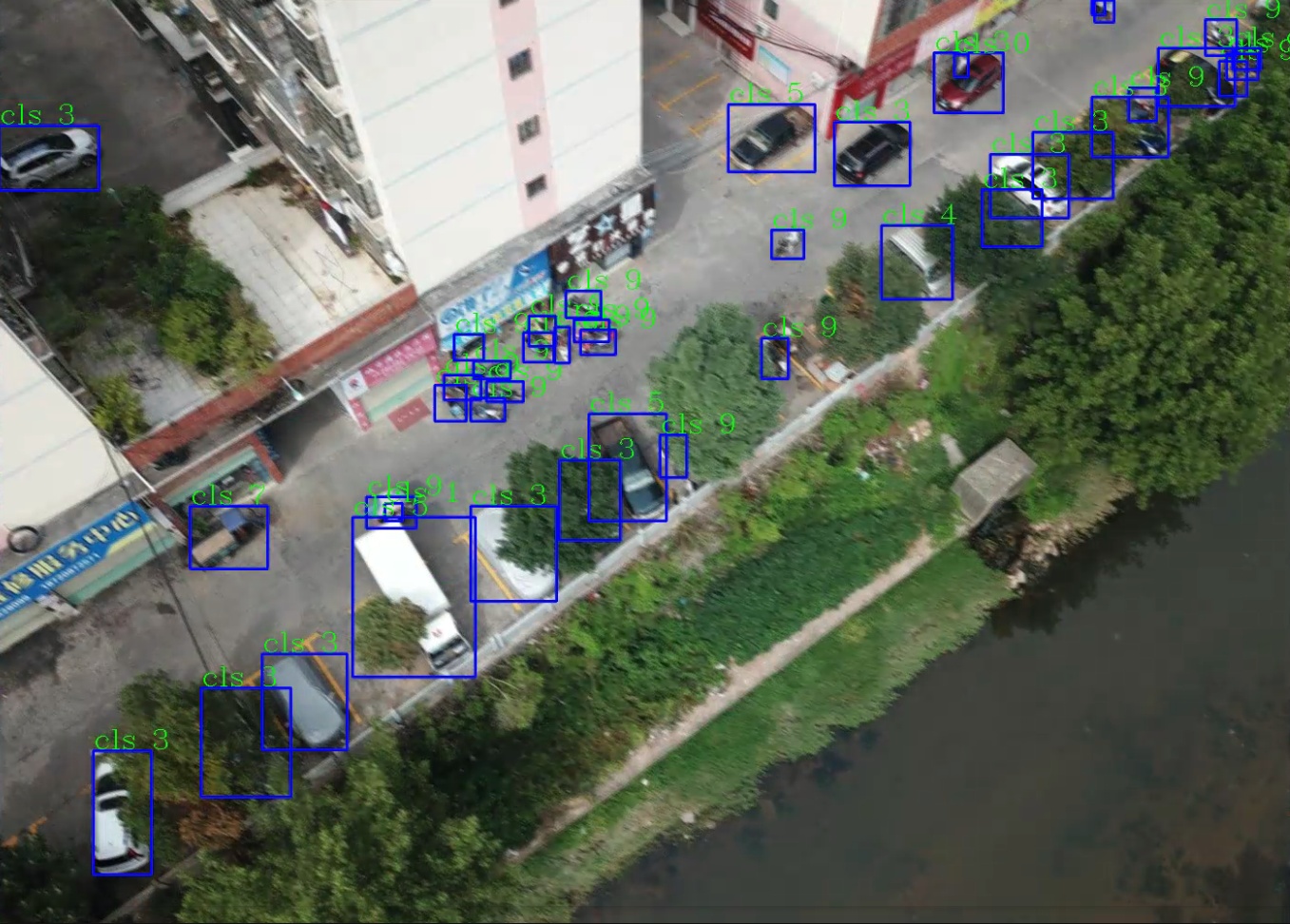}\vspace{3pt}
		
		\includegraphics[width=0.98\linewidth,height=0.63\linewidth]{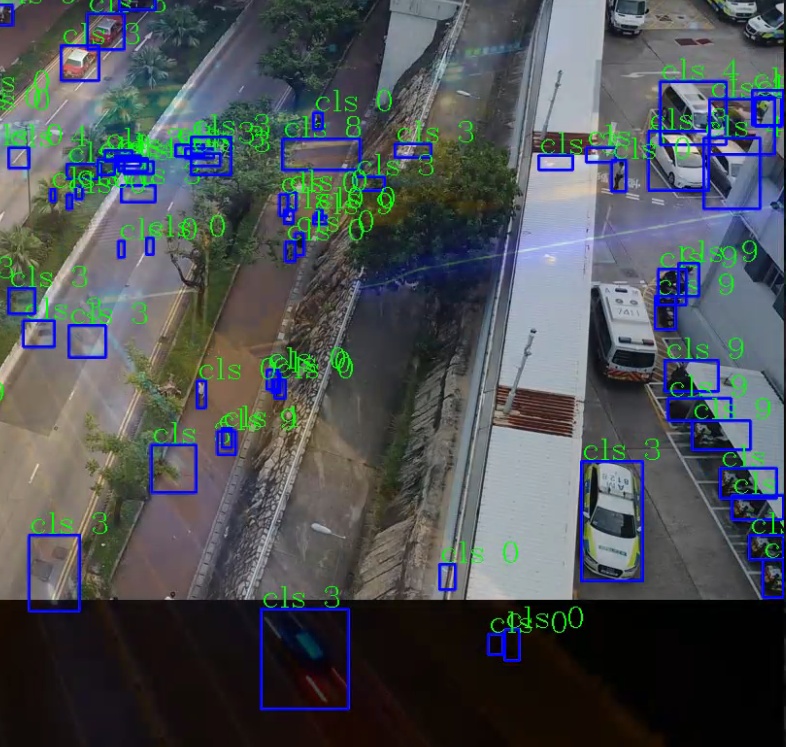}\vspace{3pt}
		
	\end{minipage}
	\caption{Visualization of the images that use random blurring and mixup data augmentation. By applying these techniques, the distribution of the training samples is significantly expanded, thus leading to stronger generalization ability of the model.}
	\label{aug_images}
\end{figure}

\begin{equation}
\label{softmax}
\alpha_i^l = \frac{\mathrm{exp}(\beta_i^l)}{\sum_{k}\mathrm{exp}(\beta_i^k)}
\end{equation}

where $\beta_i^l $ is a learnable scalar variable, which is initialized to one and, can be adaptively learned by the network through standard back-propagation.

With our proposed feature selection module, the features at different levels are aggregated adaptively. The output $Y$ is then used for the subsequent object detection following the same pipeline of CenterNet.

\section{Experiments}
\label{exps}

\begin{table}[tbp]
	\centering
	\renewcommand\tabcolsep{2pt} 
	\caption{The Comparison with state-of-the-arts on the VisDrone-DET \textit{validation} subset. All models are trained on VisDrone-DET train set. $\dag$ implies that the multi-scale inference is employed.}
	\label{MRes_VisD_val}
	\begin{tabular}{c|c|ccc|cccc}
		\toprule
		Method        & Backbone       & AP             & AP$^{50}$           & AP$^{75}$           & AR$^1$           & AR$^{10}$          & AR$^{100}$   & AR$^{500}$      \\ \midrule
		Cascade R-CNN~\cite{cai2018cascade} & ResNet50       & 24.10          & 42.90          & 23.60          & 0.40          & 2.30          & 21.00          & 35.20          \\
		Faster R-CNN~\cite{ren2015faster}  & ResNet50       & 23.50          & 43.70          & 22.20          & 0.34          & 2.20          & 18.30          & 35.70          \\ 
		RetinaNet~\cite{lin2017focal}     & ResNet50       & 15.10          & 27.70          & 14.30          & 0.17          & 1.30          & 24.60          & 25.80          \\
		FCOS~\cite{tian2019fcos}          & ResNet50       & 16.60          & 28.80          & 16.70          & 0.38          & 2.20          & 24.40          & 24.40          \\
		\midrule[0.75pt]
		HFEA~\cite{zhang2019fully}          & ResNeXt101     & 27.10          &  -              &        -        &               &               &                &                \\
		HFEA~\cite{zhang2019fully}          & ResNeXt152     & 30.30          &  -              &         -       &               &               &                &                \\
		ClusDet~\cite{yang2019clustered}       & ResNeXt101     & 32.40          & 56.20          & 31.60          &               &               &                &                \\
		CPEN+FPEN~\cite{tang2020penet}     & Hourglass104   & 36.70          & 53.20          & \textbf{39.50} &               &               &                &                \\
		DMNet+DSHNet~\cite{yu2020towards}  & ResNet50       & 30.30          & 51.80          & 30.90          &               &               &                &                \\
		DSOD~\cite{zhang2019dense}          & ResNet50       & 28.80          & 47.10          & 29.30          &               &               &                &                \\
		HRDNet~\cite{liu2020hrdnet}        & ResNet18+101   & 28.33          & 49.25          & 28.16          & 0.47          & 3.34          & 36.91          & 36.91          \\
		HRDNet$\dag$~\cite{liu2020hrdnet}       & ResNeXt50+101  & 35.51          & 62.00          & 35.13          & 0.39          & 3.38          & 30.91          & 46.62          \\
		\midrule[0.75pt]
		Ours     & CBResNet50     & 33.95          & 60.46          & 32.69          & 0.66          & 7.25          & 40.28          & 49.84          \\
		Ours$\dag$     & CBResNet50     & 37.62          & 65.41          & 37.06          & 0.62          & 7.21          & 43.21          & 56.26          \\
		Ours     & CBResNet50+DCN & 35.43          & 61.88          & 34.60          & 0.80          & 7.77          & 41.77          & 51.48          \\
		Ours$\dag$     & CBResNet50+DCN & \textbf{39.48} & \textbf{66.98} & 39.45          & \textbf{0.82} & \textbf{7.94} & \textbf{44.76} & \textbf{58.46} \\ \bottomrule
	\end{tabular}
\end{table}

\subsection{Datasets}
We select two challenging datasets to validate the effectiveness of our proposed algorithm, VisDrone \cite{zhu2020vision} and PASCAL-VOC \cite{pascal_voc}. 

\noindent\textbf{VisDrone} VisDrone is a very challenging dataset, with a wide range of variation in scale, lighting condition, scenario, viewpoint, and blurring. This dataset contains a total of 10209 images for object detection on drones, which are captured by various drone platforms, and from 14 different cities in China, including various scenarios. Ten object categories, \eg, car and van, are considered for evaluation. Specifically, 10209 images are split into four subsets, including 6471 images in the \textit{training} subset, 548 in the \textit{validation} subset, 1610 in the \textit{test-dev} subset, and 1580 in the \textit{test-challenge} subset. 

\noindent\textbf{PASCAL-VOC} VOC is a popular dataset for object detection. The union set of 2007 and 2012 train\&val is used for training, including 16551 images. The evaluation is performed on the VOC 2007 test set, with 4962 testing images, and a total of 20 categories.

\subsection{Evaluation Metrics} 
For the VisDrone dataset, we use the same evaluation protocol proposed in \cite{zhu2020vision}, which is almost the same as COCO \cite{lin2014microsoft}, except for reporting an additional $\mathrm{AR^{max=500}}$ score. The mean average precision (mAP) over thresholds of $\mathrm{IoU}=0.5:0.05:0.95$ is reported. Moreover, we report the $\mathrm{AP^{IoU=0.5}}$, $\mathrm{AP^{IoU=0.75}}$, $\mathrm{AR^{max=1}}$, $\mathrm{AR^{max=10}}$, and $\mathrm{AR^{max=100}}$ scores. For the VOC dataset, we use mAP at IOU threshold 0.5 as the evaluation metric.

\begin{table}[tbp]
	\centering
	\renewcommand\tabcolsep{7pt} 
	\caption{Comparison results of various algorithms on the VisDrone-DET \textit{test-challenge} subset. $\dag$ implies that the multi-scale inference is employed.}
	\label{MRes_VisD_test}
	\begin{threeparttable}
		\begin{tabular}{c|cccc|ccc}
			\toprule
			Method         & $\mathrm{AP}$ & $\mathrm{AP}^{50}$ & $\mathrm{AP}^{75}$ & $\mathrm{AR}^1$ & $\mathrm{AR}^{10}$ & $\mathrm{AR}^{100}$ & $\mathrm{AR}^{500}$ \\ \midrule
			DPNet-ensemble & 29.62 & 54.00 & 28.70 & 0.58          & 3.69          & 17.10          & 42.37          \\
			RRNet          & 29.13 & 55.82          & 27.23          & 1.02 & 8.50 & {35.19} & 46.05 \\
			ACM-OD         & 29.13 & 54.07          & 27.38          & 0.32          & 1.48          & 9.46           & 44.53          \\
			S+D            & 28.59 & 50.97          & 28.29          & 0.50          & 3.38          & 15.95          & 42.72          \\
			BetterFPN      & 28.55 & 53.63          & 26.68          & 0.86          & 7.56          & 33.81          & 44.02          \\
			HRDet+         & 28.39 & 54.53          & 26.06          & 0.11          & 0.94          & 12.95          & 43.34          \\
			CN-DhVaSa      & 27.83 & 50.73          & 26.77          & 0.00          & 0.18          & 7.78           & 46.81     \\
			MSC-CenterNet  & 31.13 & 54.13 & 31.41 & 0.27 & 1.85 & 6.12  & 50.48 \\
			CenterNet+     & 30.94 & 52.82 & 31.13 & 0.27 & 1.84 & 5.67  & 50.93 \\
			PG-YOLO        & 26.05 & 49.63 & 24.15 & \textbf{1.45} & 9.20 & 33.65 & 42.63 \\ 
			Ours$\dag$           & \textbf{32.34} & \textbf{56.46} & \textbf{32.39} & 1.20 & \textbf{9.45} & \textbf{38.55} & \textbf{51.61} \\
			\bottomrule
		\end{tabular}
	\end{threeparttable}
\end{table}

\subsection{Implementation Details}
\noindent\textbf{Training} We implement our model on the deep-learning framework of PaddlePaddle \cite{ma2019paddlepaddle}. For the VisDrone dataset, the input resolution of the network is set to $1023 \times 767$ during training, which results in an output resolution of $256 \times 192$. For the VOC dataset, the training resolution is fixed at $512\times 512$. We use extensive data augmentation techniques to reduce the risk of overfitting. For standard data augmentation, we adopt random cropping, random scaling, random flipping, random blurring, and random color jittering. Moreover, we use the mixup algorithm \cite{zhang2017mixup} to improve the generability of the model. Some example images of our data augmentation techniques are showed in Fig. \ref{aug_images}.

Adam \cite{kingma2014adam} algorithm is adopted to optimize the overall training objective. We train our model on a V100 GPU with a batch size of 8. Note that for ablation study, we train our model for 50k iterations, with an input resolution of $511 \times 511$, to save computational resources. When comparing to other state-of-the-arts, we train the model for a total of 120k iterations. The learning rate starts from 2e-4 and, is decayed by cosine annealing strategy. 

\noindent\textbf{Inference} Following the setting of CenterNet, we use three levels of test augmentation, \ie, no augmentation, flipping, and multi-scale (1.0, 1.25, 1.5, and 1.8) test augmentation. Specifically, for flipping test augmentation, the network predictions of the flipped and non-flipped inputs are average before decoding. For multi-scale test augmentation, non-maximum suppression (NMS) is further applied to filter out the duplicate predictions.

\begin{table}[tbp]
	\centering
	\renewcommand\tabcolsep{8pt} 
	\caption{The performance comparison on  VOC 2007 test set, in terms of mAP (\%).}
	\label{MRes_Voc_07}
	\begin{tabular}{c|c|c|c|c}
		\toprule
		Method       & Backbone      & Input size & mAP            & FPS   \\ \midrule
		Faster R-CNN~\cite{ren2015faster} & ResNet101     & 1000$\times$600   & 76.4           & 5     \\
		R-FCN~\cite{dai2016rfcn}     & ResNet101     & 1000$\times$600   & 80.5           & 9     \\ 
		OHEM~\cite{shrivastava2016training}  & VGG-16        & 1000$\times$600   & 74.6           & -    \\
		R-FCN~\cite{dai2016rfcn}  & ResNet101+DCN     & 1000$\times$600   & 82.6           & -    \\
		CoupleNet~\cite{zhu2017couplenet}  & ResNet101     & 1000$\times$600   & 82.7           & 8.2   \\
		DeNet(wide)~\cite{ghodrati2015deepproposal}  & ResNet101     & 512$\times$512    & 77.1           & 33    \\
		FPN-Reconfig~\cite{kong2018deep} & ResNet101     & 1000$\times$600   & 82.4           &  -    \\ \midrule
		Yolov2~\cite{redmon2017yolo9000}       & DarkNet19     & 544$\times$544    & 78.6           & 40    \\
		SSD~\cite{liu2016ssd}          & VGG-16        & 513$\times$513    & 78.9           & 19    \\
		DSSD~\cite{fu2017dssd}         & VGG-16        & 513$\times$513    & 81.5           & 5.5   \\
		RefineDet~\cite{zhang2018single}    & VGG-16        & 512$\times$512    & 81.8           & 24    \\
		CenterNet~\cite{zhou2019objects}    & ResNet101     & 512$\times$512    & 78.7           & 30    \\
		CenterNet~\cite{zhou2019objects}    & DLA           & 512$\times$512    & 80.7           & 33    \\
		HRDNet~\cite{liu2020hrdnet}       & ResNeXt50+101 & 2000$\times$1200  & 82.4           &  -    \\ \midrule
		Ours         & CBResNet50    & 511$\times$511    & 78.2           & 25.08 \\
		Ours     & CBResNet50+DCN    & 511$\times$511    & 81.05          & 21.98 \\
		Ours     & ResNet101+DCN   & 511$\times$511    & \textbf{83.04} & 15.96 \\ \bottomrule
	\end{tabular}
\end{table}

\subsection{Comparison with state-of-the-arts}
We first demonstrate the advancement of our proposed method on the very challenging VisDrone dataset. Then we move on to prove the strong generality of our model on the VOC dataset.

\noindent\textbf{VisDrone} The comparison results with other state-of-the-art methods on VisDrone2019 DET \textit{test-challenge} and \textit{validation} subset is showed in Table \ref{MRes_VisD_test} and Table \ref{MRes_VisD_val}, respectively. Our model achieves the best performance among other detectors, with an AP score of 32.34\% on the \textit{test-challenge} subset, and 39.48\% AP on the \textit{validation} subset, respectively. This outperforms other algorithms by a large margin, with an improvement of 7.08\% AP over ClusDet \cite{yang2019clustered}, demonstrating the effectiveness of our proposed model. It is worth noting that the average recall (AR) obtained by the proposed model also outperforms other methods at all levels (from AR$^1$ to AR$^{500}$), which shows that our model can better cope with the variation in object scale (find out more object instance), with the help of adaptive feature selection module.

\noindent\textbf{PASCAL-VOC} Table \ref{MRes_Voc_07} tabulates the results of various state-of-the-art methods on VOC dataset. Note that flipping test augmentation is used by default. Our method achieves the best performance, with an mAP score of 83.04\%, while still running in real-time (15.96 FPS). DSSD yields 81.5\% mAP, which is comparable to our method using CBResNet50 \cite{liu2020cbnet} as the backbone, whereas the inference speed is approximately one third of ours (5.5 FPS \textit{vs.} 21.98 FPS). Therefore, our method performs well on the balance of speed and accuracy. 

\begin{figure}[tbp]
	\centering
	\begin{minipage}[t]{0.31\linewidth}
		\includegraphics[width=0.98\linewidth,height=0.55\linewidth]{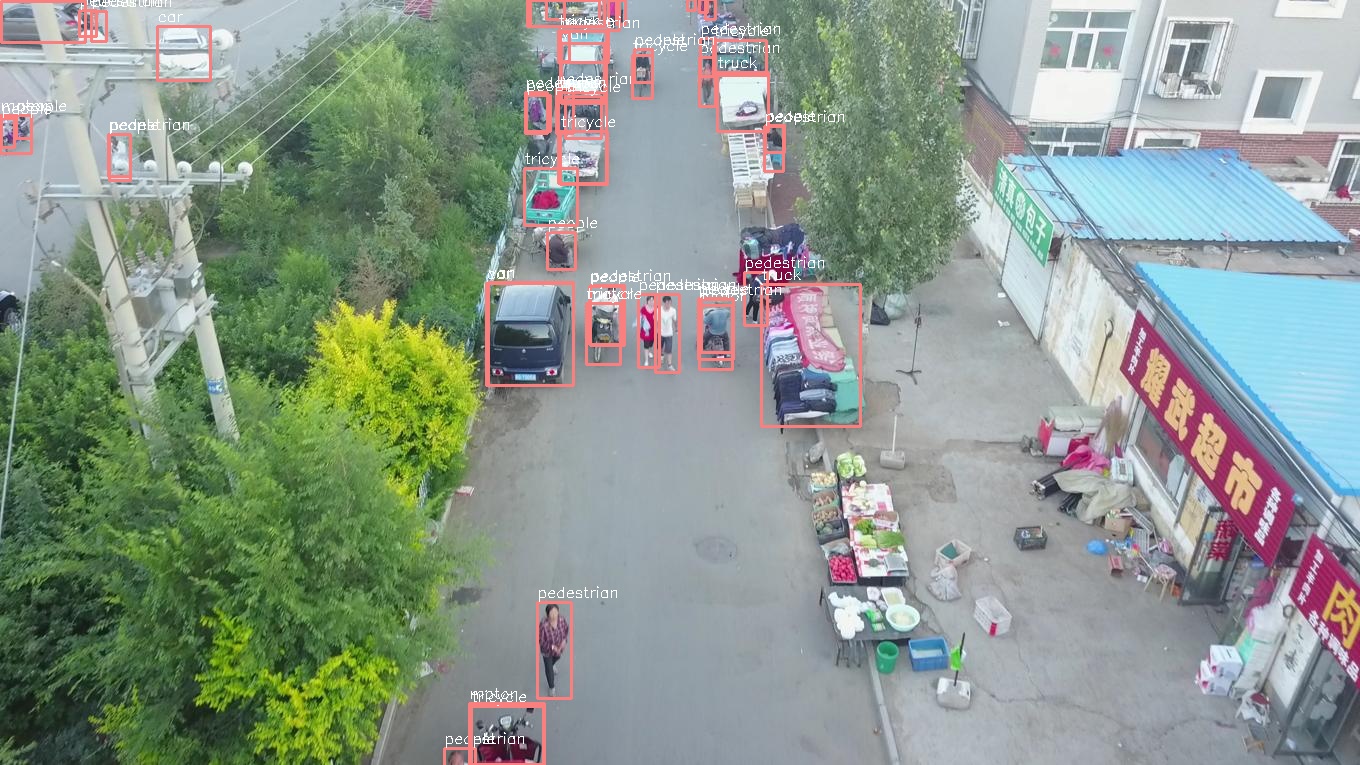}\vspace{3pt}
		\includegraphics[width=0.98\linewidth,height=0.55\linewidth]{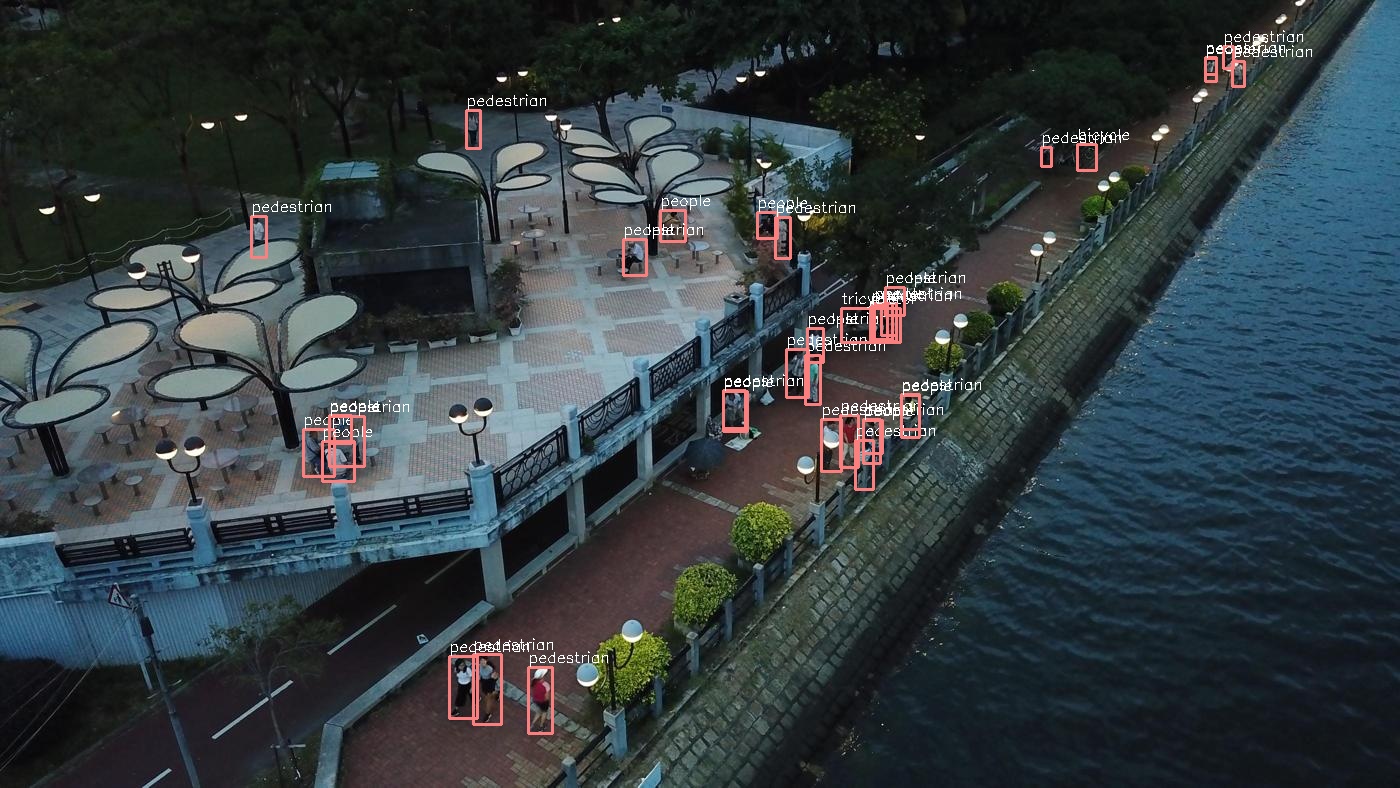}\vspace{3pt}
		\includegraphics[width=0.98\linewidth,height=0.55\linewidth]{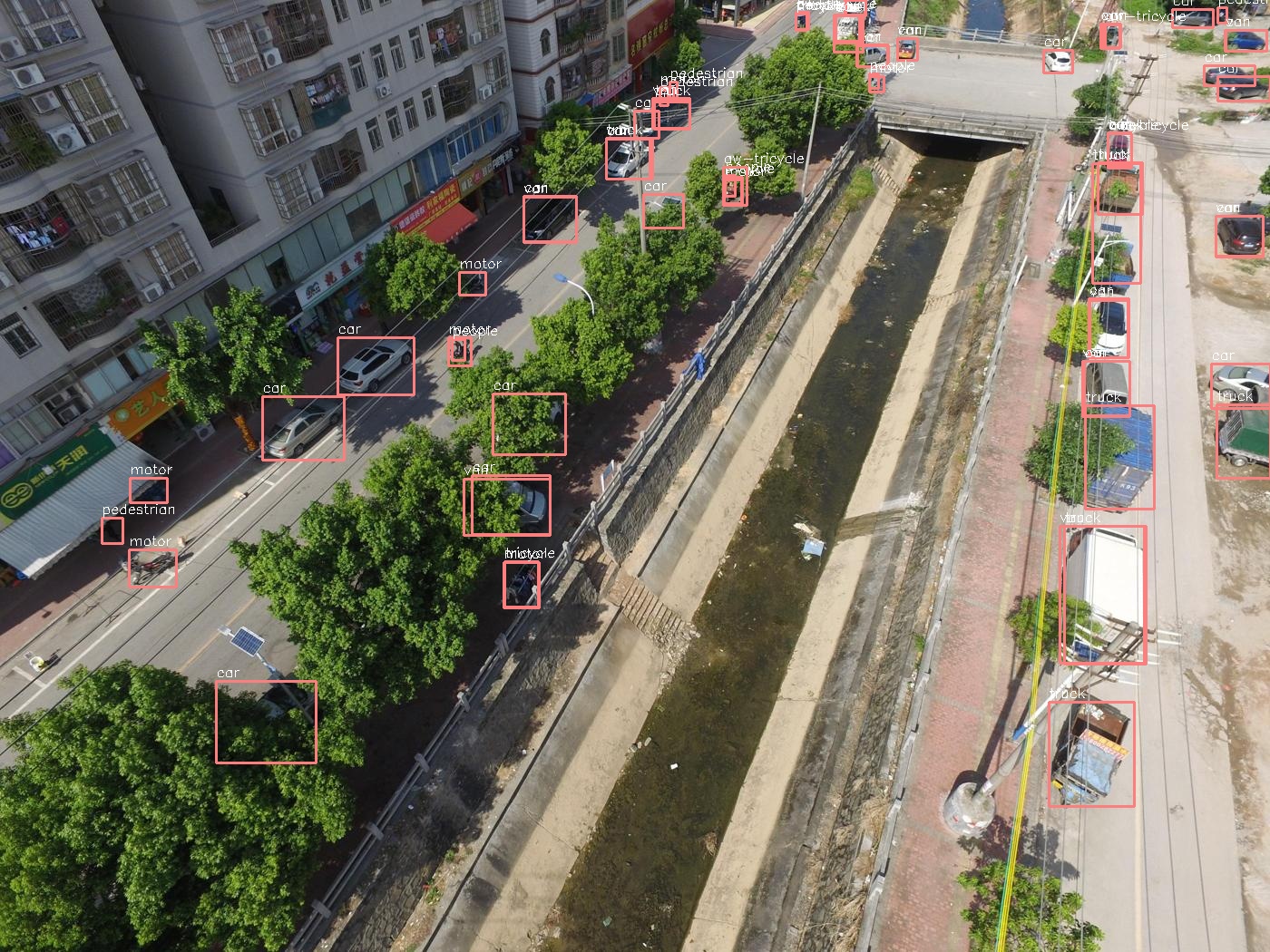}\vspace{3pt}
	\end{minipage}
	\begin{minipage}[t]{0.31\linewidth}
		\includegraphics[width=0.98\linewidth,height=0.55\linewidth]{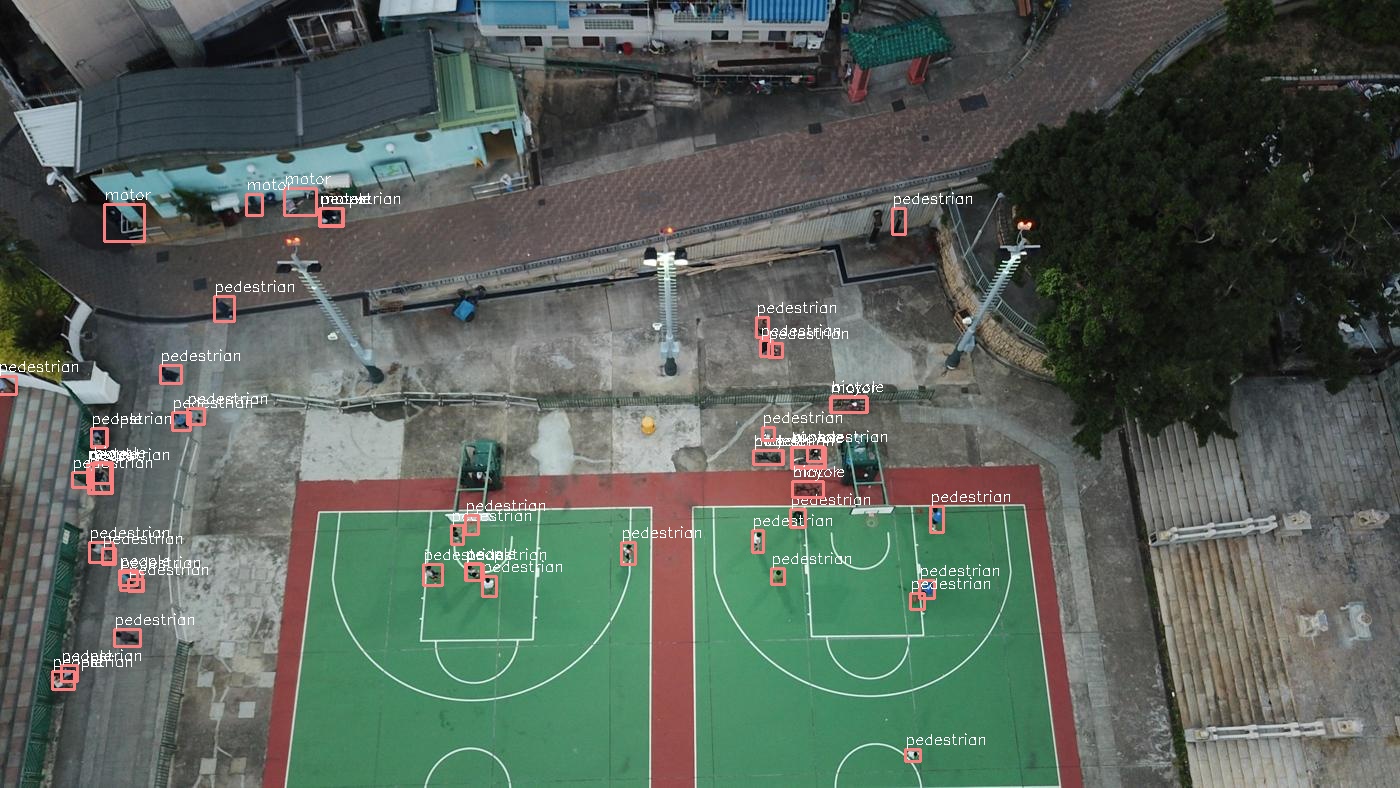}\vspace{3pt}
		\includegraphics[width=0.98\linewidth,height=0.55\linewidth]{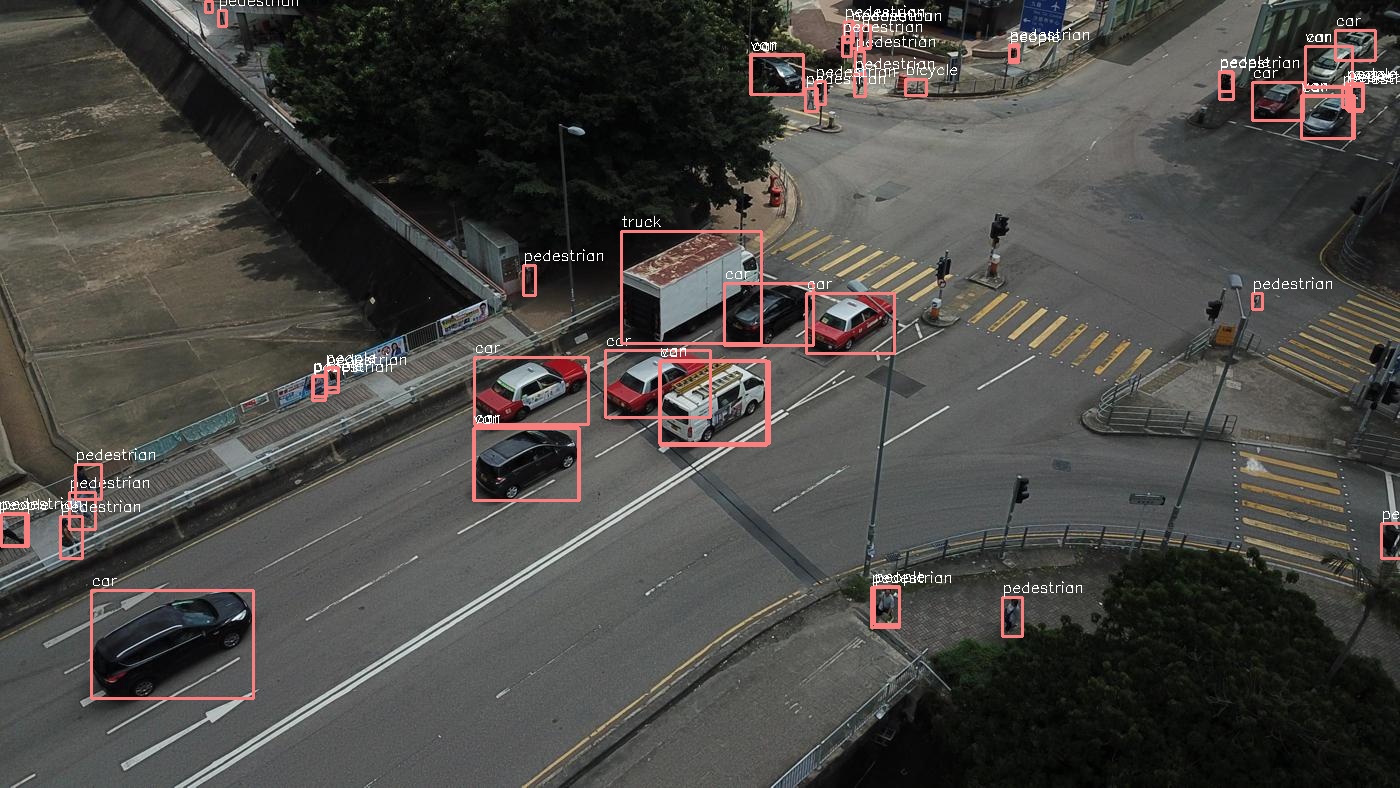}\vspace{3pt}
		\includegraphics[width=0.98\linewidth,height=0.55\linewidth]{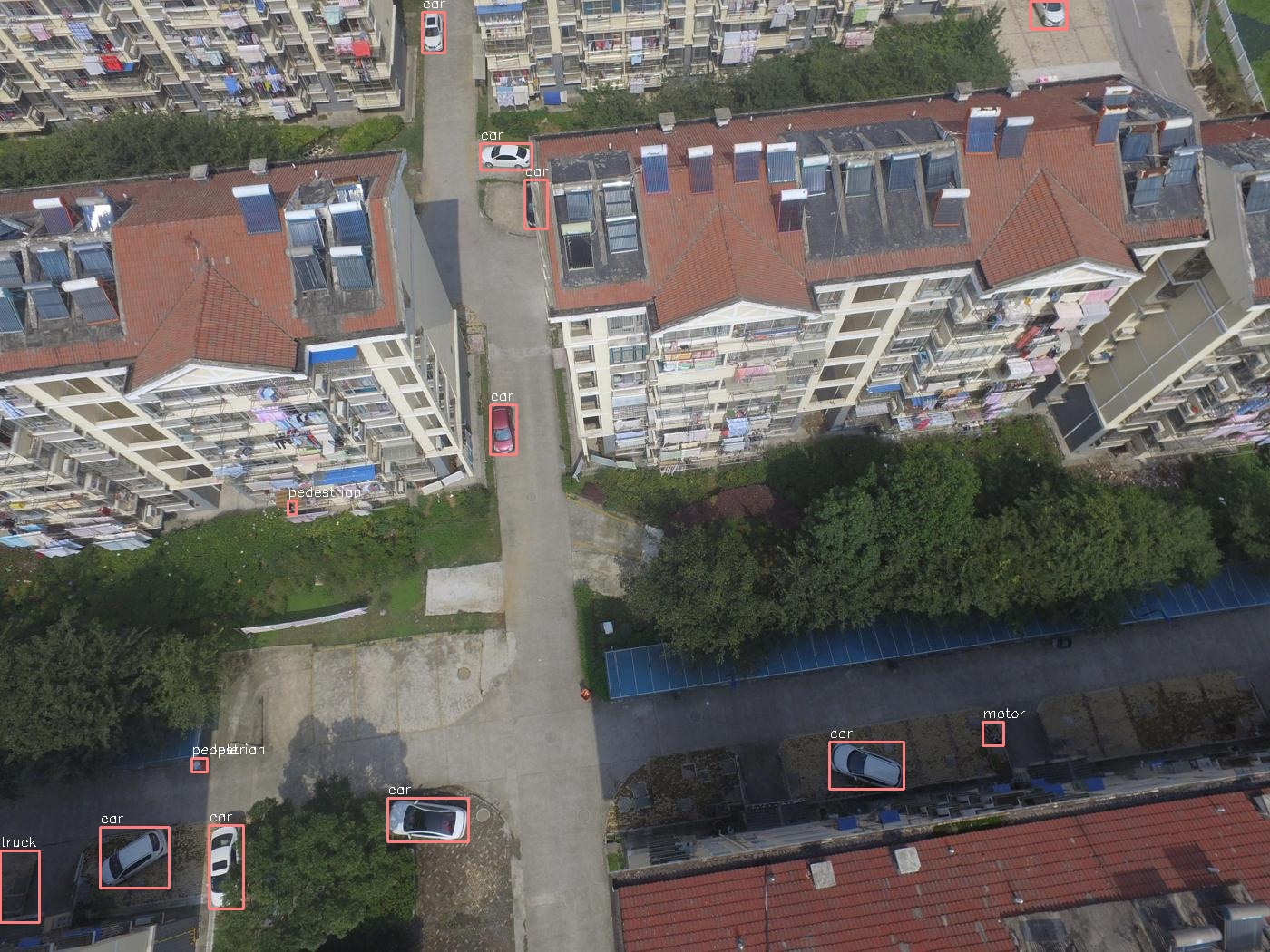}\vspace{3pt}
	\end{minipage}
	\begin{minipage}[t]{0.31\linewidth}
		\includegraphics[width=0.98\linewidth,height=0.55\linewidth]{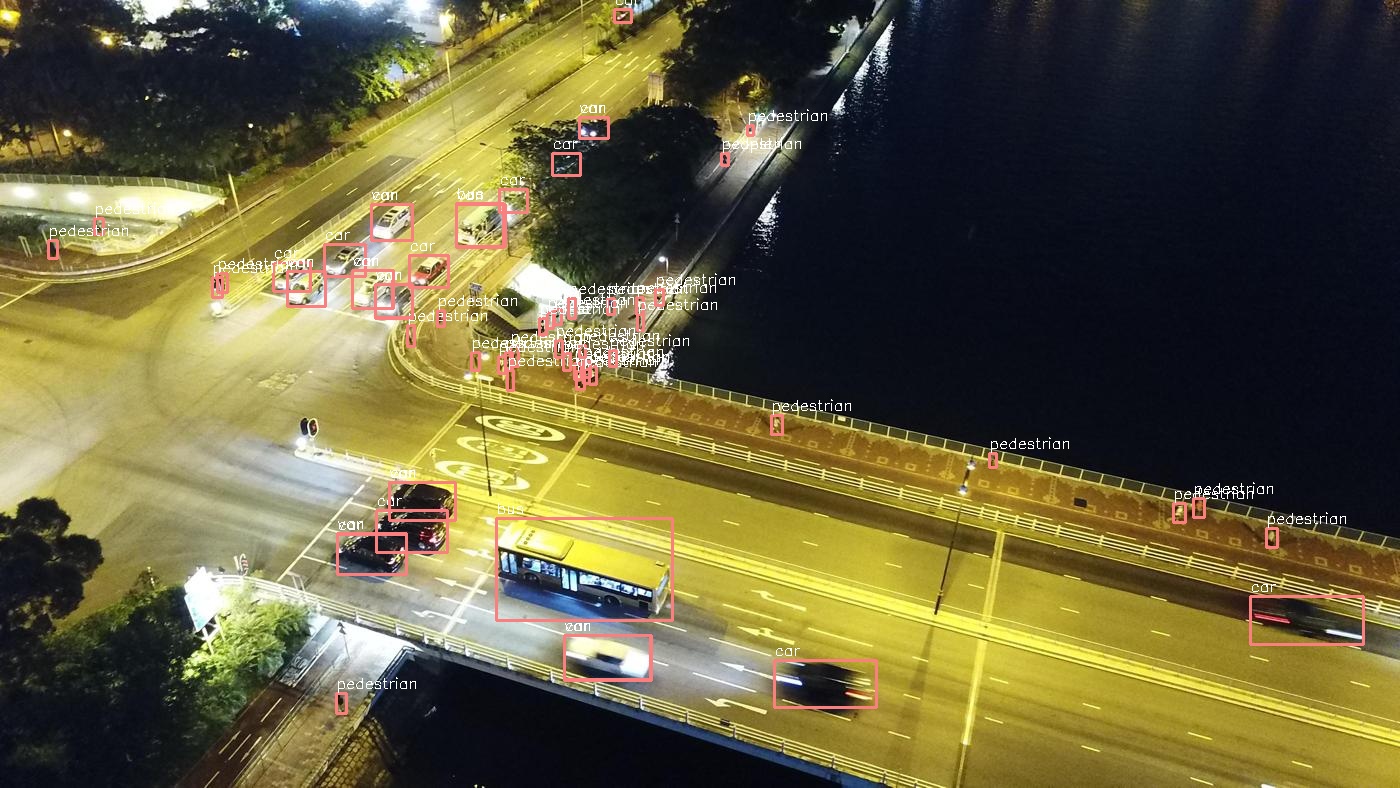}\vspace{3pt}
		\includegraphics[width=0.98\linewidth,height=0.55\linewidth]{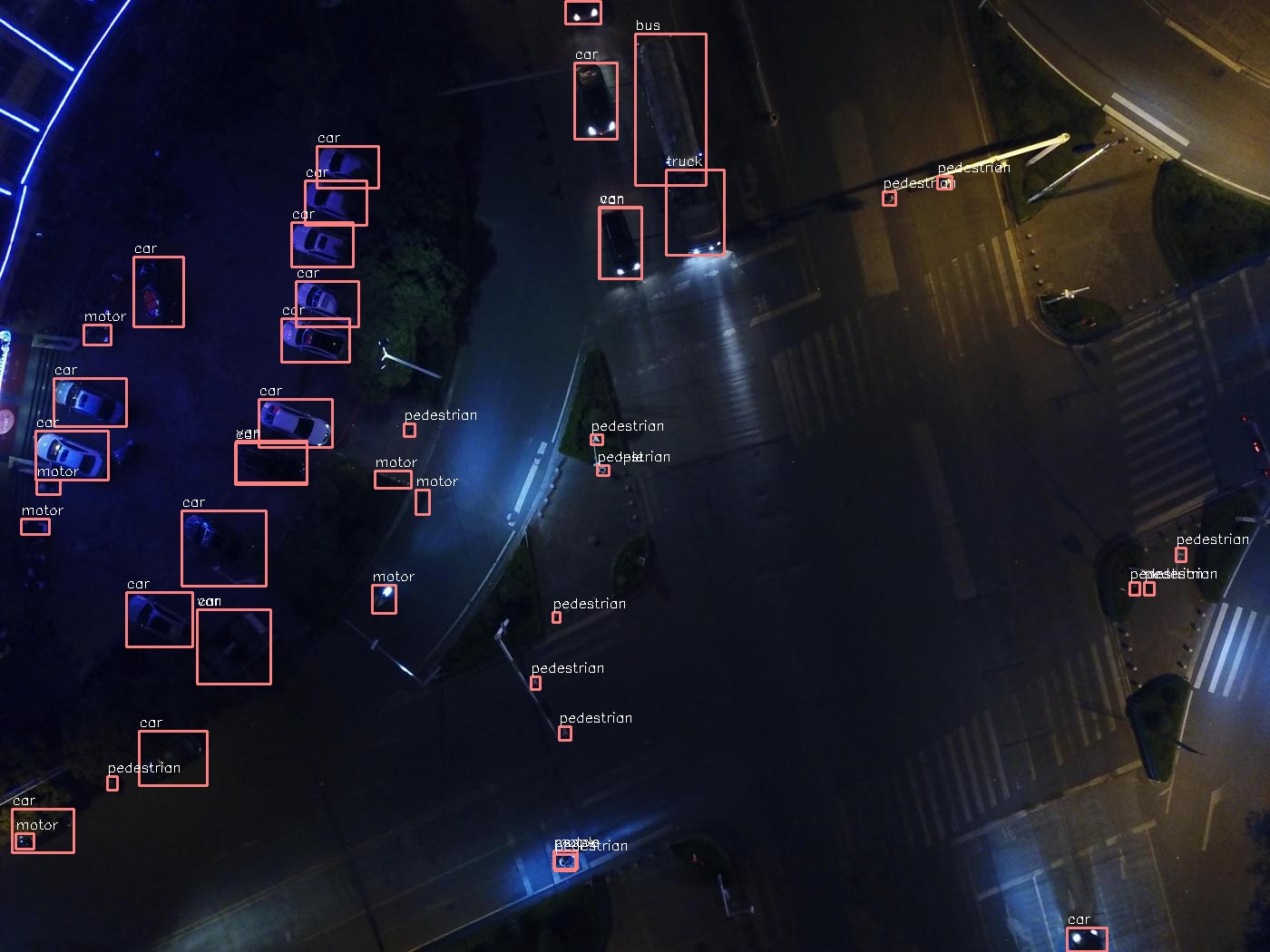}\vspace{3pt}
		\includegraphics[width=0.98\linewidth,height=0.55\linewidth]{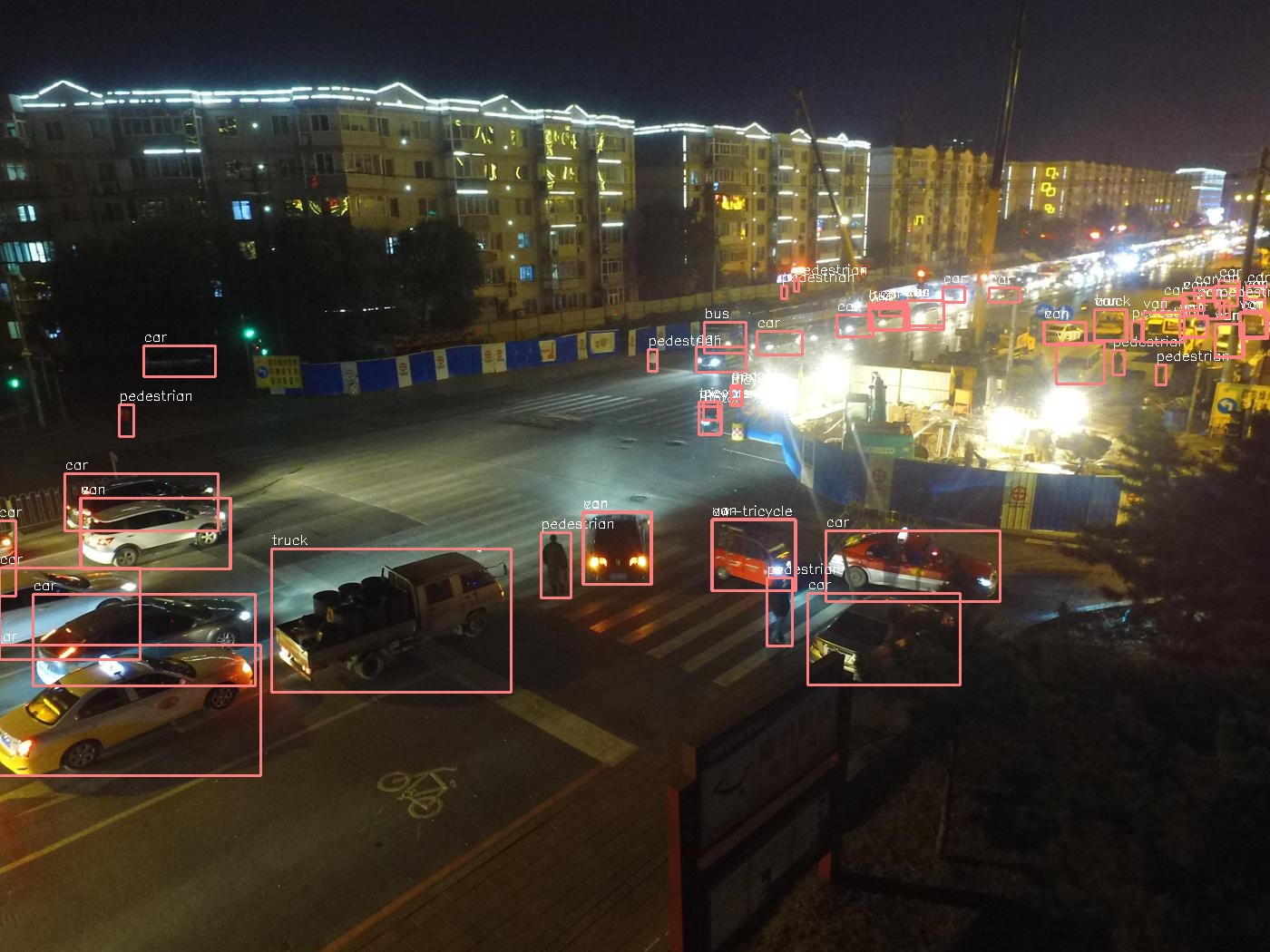}\vspace{3pt}
	\end{minipage}
	\caption{Qualitative visualization of the detection results on VisDrone-DET \textit{test-challenge} subset. Our model is capable of dealing with objects with a wide range of variation in shape, viewpoint, lighting condition, and blurring, etc.}
	\label{vis_det_results}
\end{figure}

\subsection{Ablation Studies}
\label{sec_abla}

\noindent\textbf{Strong baseline}
The contribution of each advanced components to our baseline detector is tabulated in Table \ref{Abla_strong_baseline}, where GCB denotes the global context block \cite{cao2019gcnet}, CASM is the class-aware sampling mechanism, and M-GIOU denotes the modified GIOU loss for object size regression. As shown in Table \ref{Abla_strong_baseline}, all the techniques boost the performance, \eg, adding CASM to the baseline method improves the AP from 25.92\% to 26.71\%, with an increase of 0.79\%. Thanks to these techniques, we produce a stronger baseline that delivers an AP score of 28.34\%, an improvement of 2.42\% over the CenterNet baseline. It is worth noting that the improvement is almost cost-free, as mixup, CASM, and M-GIOU do not introduce any extra computational cost, and GCB owns the light-weight property. Therefore, the final baseline still runs in an FPS of 20.76 on a single GPU of NVIDIA Tesla V100.

\noindent\textbf{Influence of the regression weight $\lambda_{size}$} Further on we study the influence of the regression weight on the detector accuracy, as shown in Table \ref{Abla_lambda}. $\lambda_{size} = 1.0$ yields good results, with an AP score of 26.85\%. For larger ($\lambda_{size}=5.0$) or smaller ($\lambda_{size}=0.2$) values, the AP degrades a little bit, with 0.23\% and 0.57\% AP, respectively. This shows that M-GIOU loss is not much sensitive to the variation of $\lambda_{size}$.

\noindent\textbf{Channel fusion \textit{vs.} Spatial fusion} To analyze the effectiveness of the proposed AFSM, we further report the results of ASFF \cite{liu2019learning} based on the CenterNet baseline. As shown in Table \ref{Abla_afsm_asff}, applying ASFF to the baseline improves the performance by 3.23\% AP, showing that fusing multi-level representations via point-wise spatial addition helps the model to tackle the variation in object scale. However, adding AFSM to the baseline boosts the performance by 4.02\% AP, with an increase of 0.79\% AP compared to ASFF. This implies that fusing features across feature levels by re-weighting the importance distribution over the channel dimension may be more suitable and effective than the spatial dimension. Moreover, instead of using several $1\times 1$ convs to generate fusion weights for a specific image, we regard the fusion weight as a learnable vector. This enables AFSM to learn the distribution of object size over the whole dataset, which makes the selection process more stable and effective.

\begin{table}[tbp]
	\centering
	\renewcommand\tabcolsep{8pt} 
	\caption{Effect of each advanced techniques on the baseline, in terms of APs (\%). All models employ ResNet50 \cite{he2016deep} as backbone network. FPS are reported on VisDrone2019 DET \textit{validation} subset, keeping the original resolution for testing.}
	\label{Abla_strong_baseline}
	\begin{tabular}{ccccc|cccc}
		\toprule
		CenterNet            & GCB & mixup & CASM & M-GIOU & AP    & AP$^{50}$  & AP$^{75}$ & FPS  \\ \midrule
		\checkmark   &   &    &    &   & 25.92 & 46.81 & 24.75 & 23.20 \\
		\checkmark & \checkmark   &  &  &  & 27.51 & 49.11 & 26.66 & 20.76 \\
		\checkmark &     & \checkmark    &   &   & 26.94 & 47.87 & 26.14 & 22.58\\
		\checkmark &     &       & \checkmark    &        & 26.71 & 47.76 & 25.71 & 22.58 \\
		\checkmark &     &       &      & \checkmark      & 26.85 & 48.04 & 26.00 & 22.58 \\
		\checkmark & \checkmark   & \checkmark     & \checkmark    & \checkmark  & 28.34 & 49.92 & 27.92 & 20.76 \\ 
		\bottomrule
	\end{tabular}
\end{table}

\begin{table}[tbp]
	\centering
	\renewcommand\tabcolsep{8pt} 
	\caption{The influence of $\lambda_{size}$ on our model.}
	\label{Abla_lambda}
	\begin{tabular}{cccccccc}
		\toprule
		$\lambda_{size}$ & AP    & AP$^{50}$    & AP$^{75}$    & AR$^1$           & AR$^{10}$    & AR$^{100}$     & AR$^{500}$     \\ \midrule
		0.2    & 26.28          & 47.74          & 25.21          & 0.62          & 7.01          & 35.76          & 35.76          \\
		1.0      & \textbf{26.85} & \textbf{48.04} & \textbf{26.00} & \textbf{0.72} & \textbf{7.03} & \textbf{36.23} & \textbf{36.23} \\
		5.0      & 26.62          & 47.18          & 25.94          & 0.58          & 6.75          & 36.05          & 36.05          \\ \bottomrule
	\end{tabular}
\end{table}

\noindent\textbf{How to obtain adaptive feature selection weights} For obtaining the weights $\alpha^l \in \mathbb{R}^d$ to select representative features across multi-scale levels, we evaluate the performance based on three kinds of methods, where $d$ is the number of channels in feature maps $X^l$. The difference between these methods is the ways to obtain $\beta^l$ in Eq. \ref{softmax}. The first is a learnable vector, which is initialized to one and, can be updated by the standard back-propagation algorithm. For the second version, we first pool the feature maps $X^l$ using global average pooling (GAP) operation, obtaining $X_p^l \in \mathbb{R}^d$. Then each of the pooled features is followed by a fully connected (FC) layer, to obtain $\beta^l$, which is given as follows:

\begin{equation}
	X_p^l = \mathrm{GAP}(X^l)
\end{equation}
\begin{equation}
	\beta^l = \mathrm{FC}(X_p^l)
\end{equation}


The third version uses GAP to pool the feature maps $X^l$. Then we concatenate the features from all levels, denoted as $X_p=\mathrm{[ } ...;(X_p^l)^\intercal;...;(X_p^m)^\intercal \mathrm{] }^\intercal \in \mathbb{R}^{md}$, following by a FC layer to obtain $\gamma=\mathrm{[ } ...;(\beta^l)^\intercal;...;(\beta^m)^\intercal \mathrm{] }^\intercal \in \mathbb{R}^{md}$, where $m$ is the number of levels. For convenience, we denote the three kinds of methods as V1, V2, V3, respectively, and the results of these methods are showed in Table \ref{Abla_selection_weights}. These three methods achieve comparable performance, with 29.94\%, 29.57\%, and 29.79\% AP for V1, V2, V3, respectively. However, V1 enjoys the advantages of the fastest inference among the three versions, with an FPS of 21.71. This shows that it is not the ways of obtaining $\beta^l$ that brings the performance gain, but the mechanism of adaptive feature selection proposed in this paper. The advantages of AFSM are two-fold. On the one hand, it aggregates representations across various feature levels, thus making full use of information on different scales. On the other hand, by applying an importance weight $\alpha^l$ when aggregating representation at level $l$, the model can obtain useful information while relegating the useless parts for the subsequent detection task.

\begin{figure}[tbp]
	\centering
	\begin{minipage}[t]{0.48\linewidth}
		\centering
		\includegraphics[width=0.95\linewidth, height=0.46\linewidth]{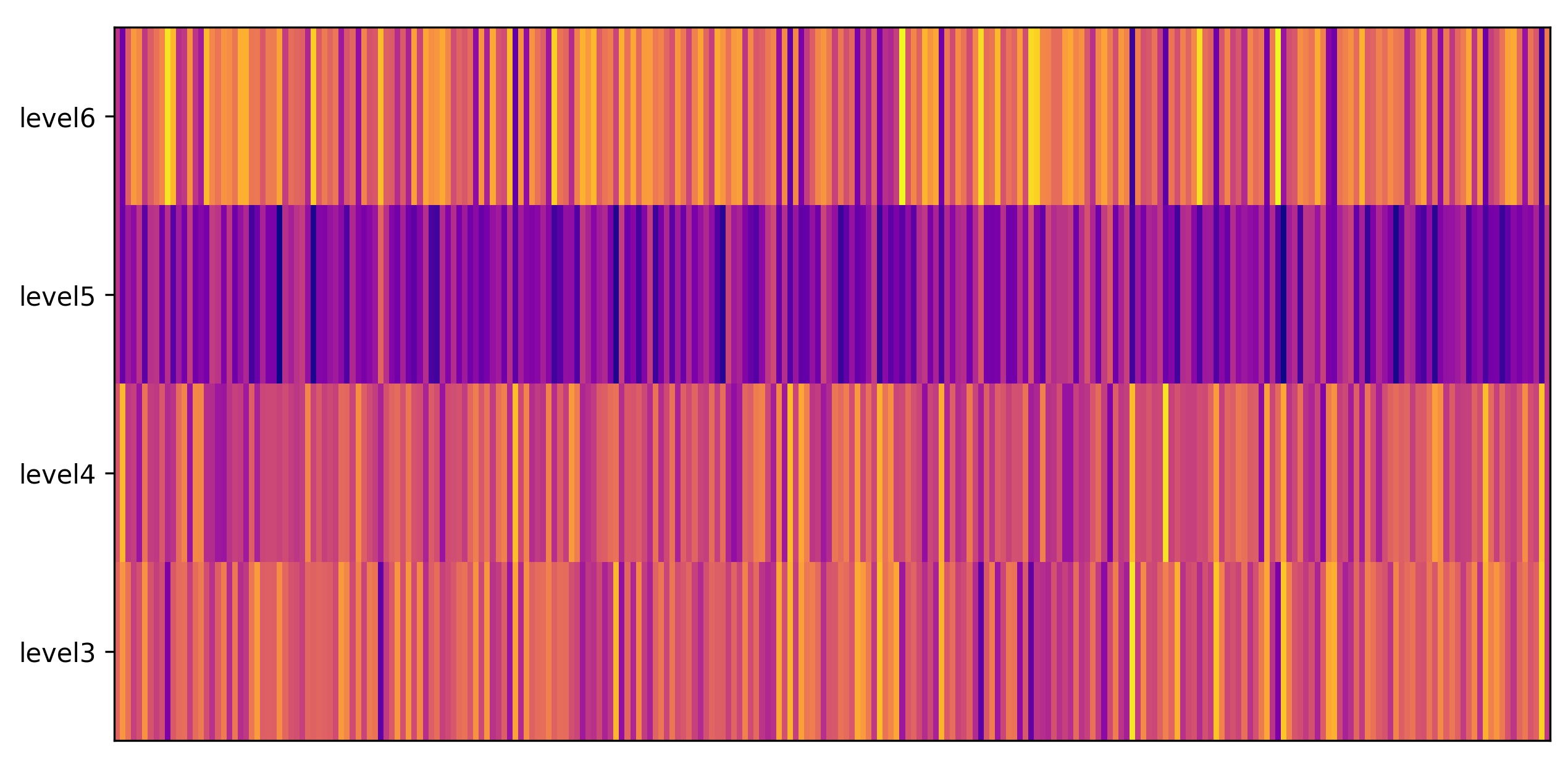}
		\subcaption{PASCAL-VOC}
		\label{voc_vis}
	\end{minipage}
	\begin{minipage}[t]{0.48\linewidth}
		\centering
		\includegraphics[width=0.95\linewidth, height=0.46\linewidth]{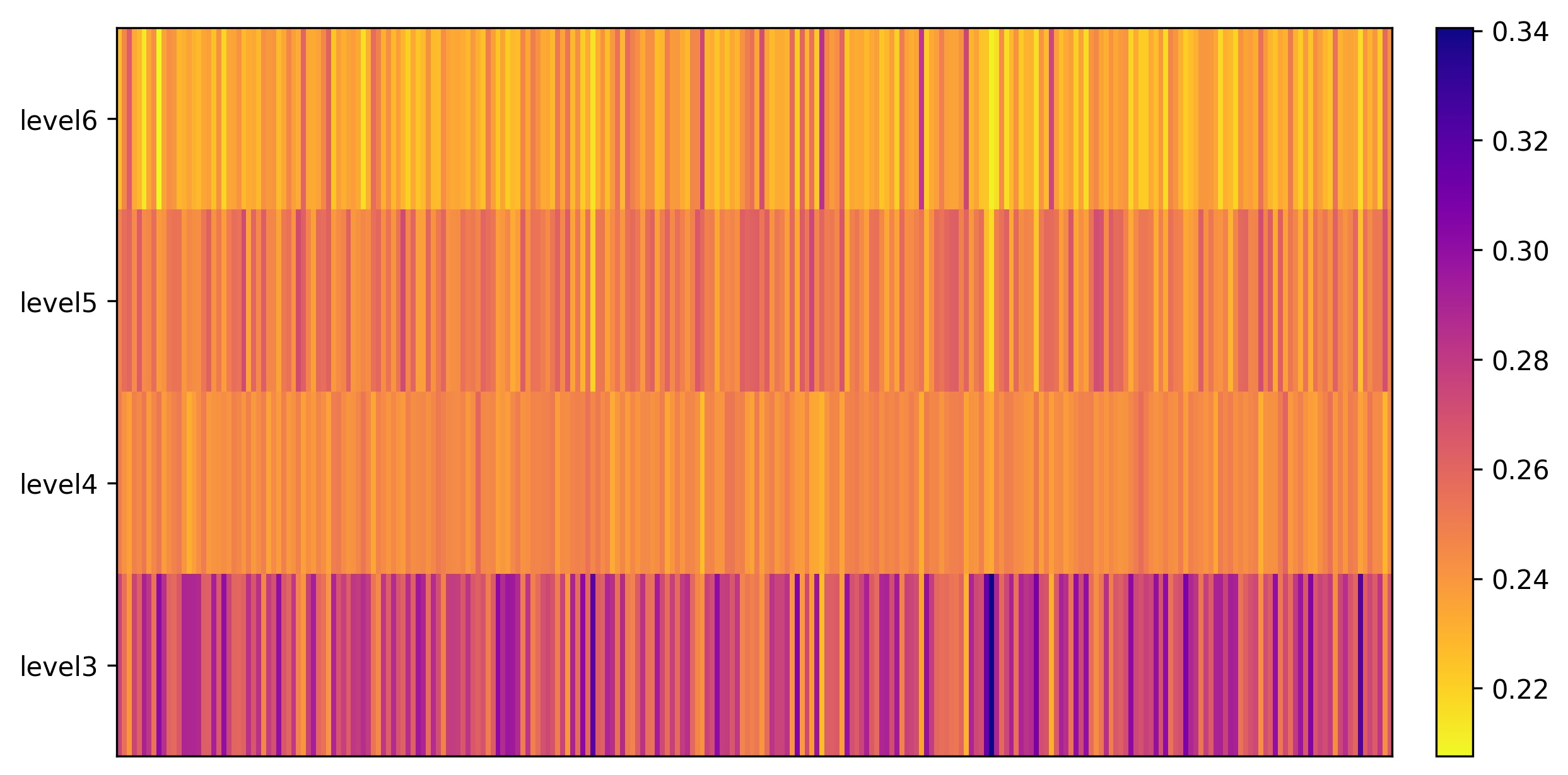}
		\subcaption{VisDrone}
		\label{visdrone_vis}
	\end{minipage}
	\caption{Visualization of the learned weights $\alpha$ at each level (from level3 to level6). The horizontal axis is the channel dimension, with 256 channels.}
	\label{vis_weights}
\end{figure}

\begin{table}[tbp]
	\centering
	\renewcommand\tabcolsep{10pt} 
	\caption{Comparison of AFSM and ASFF \cite{liu2019learning}. APs are reported on the VisDrone-DET 2019 \textit{validation} subset.}
	\label{Abla_afsm_asff}
	\begin{tabular}{ccccc}
		\toprule
		Method & AP & AP$^{50}$    & AP$^{75}$ & FPS  \\ \midrule
		Baseline & 25.92 & 46.81 & 24.75 & 23.20  \\
		+ ASFF  & 29.15 & 54.04 & 27.34 & 22.34 \\
		+ AFSM     & \textbf{29.94} & 55.40  & 28.28 & 22.76 \\ \bottomrule
	\end{tabular}
\end{table}

\begin{table}[tbp]
	\centering
	\renewcommand\tabcolsep{10pt} 
	\caption{Different ways of obtaining feature selection weights. The APs are reported on the VisDrone-DET \textit{validation} subset. Baseline is CenterNet with ResNet50 as backbone network, and uses the feature maps at level 3 for the following detection head. Three kinds of methods yield comparable performance, while V1 achieving the fastest inference.}
	\label{Abla_selection_weights}
	\begin{tabular}{ccccc}
		\toprule
		Method & AP              & AP$^{50}$    & AP$^{75}$    & FPS            \\ \midrule
		Baseline &25.92& 46.81 & 24.75& \textbf{23.20} \\
		+ V1     & \textbf{29.94} & \textbf{55.4} & \textbf{28.28} & {22.76} \\
		+ V2     & 29.57          & 55.3          & 27.58          & 17.77          \\
		+ V3     & 29.79          & 55.15         & 28.04          & 18.02          \\ \bottomrule
	\end{tabular}
\end{table}

\noindent\textbf{Visualization of feature selection weights} In order to analyze whether the proposed adaptive feature selection module learns to aggregate useful information from different feature levels, we visualize the learned weights $\alpha$ at each level as a heatmap. As shown in Fig. \ref{vis_weights}, the selection weight $\alpha^l \in \mathbb{R}^{d}$ at level $l$ is exhibited horizontally, where $d=256$ in this paper. It can be observed from Fig. \ref{vis_weights} that, AFSM successfully captures the scale distribution of the specific dataset, and learns to adaptively fuse features across multi-scale levels. Specifically, for the VOC dataset, the object scale is relatively uniform and, there is not so much small object. Therefore, the semantic information at high scale level, \eg, level 5, plays a more important role for predicting the large objects, as shown in Fig. \ref{voc_vis}. On the other hand, the object scale varies significantly on the VisDrone dataset and, most of them are small objects. It is hence the spatially finer texture information is much beneficial for detecting the vast majority of small objects (see Fig. \ref{visdrone_vis}). 

\begin{figure}[tp]
	\centering
	\includegraphics[width=0.95\linewidth]{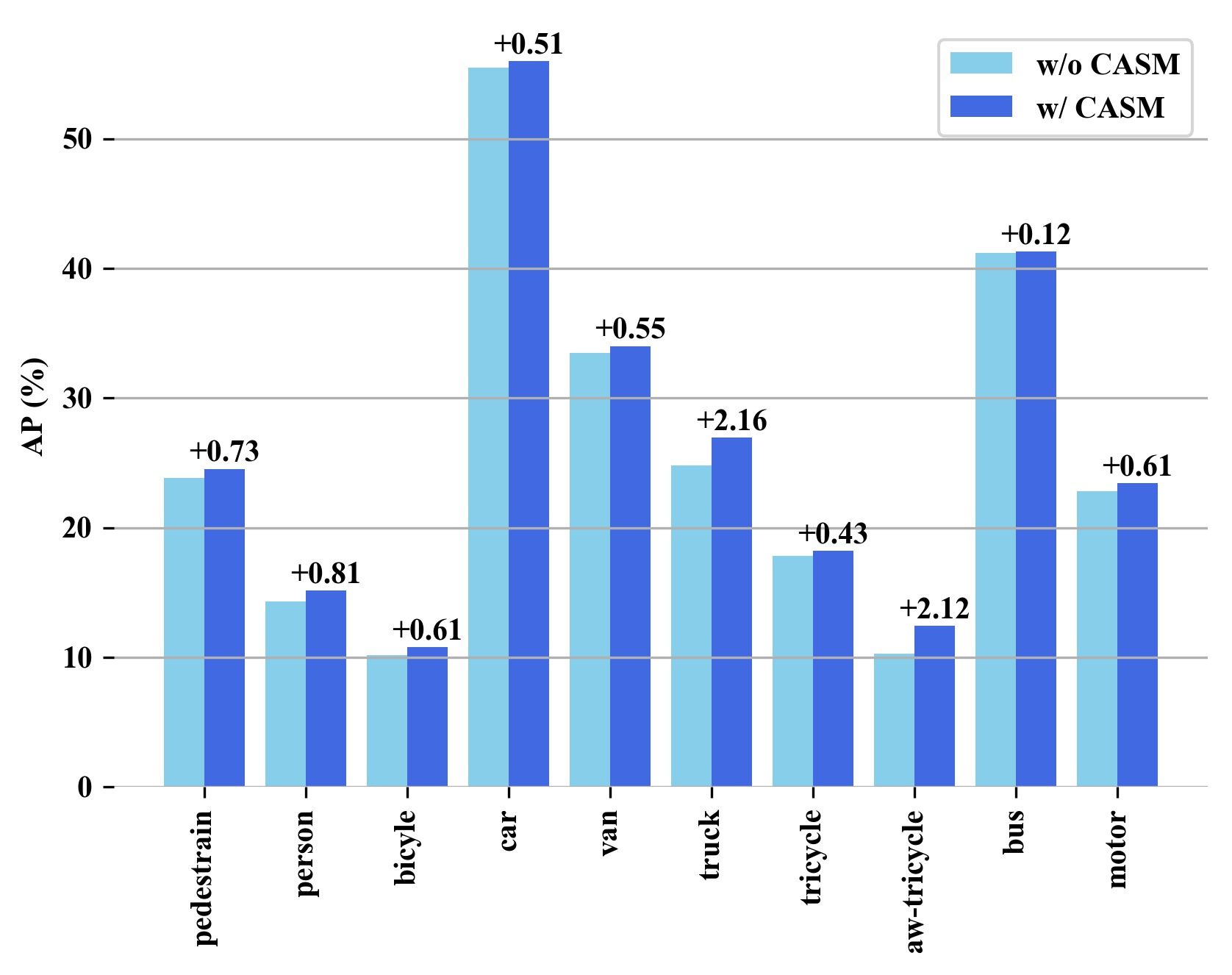}
	\caption{Effect of the class-aware sampling mechanism (CASM). CASM boosts the performance for each of classes, especially for `awning tricycle', with an improvement of 2.12\% AP.}
	\label{Abla_casm}
\end{figure}

\noindent\textbf{Effectiveness of the class-aware sampling mechanism} To analyze the effect of the proposed class-aware sampling mechanism, we plot the AP (\%) of each class on the VisDrone-DET \textit{validation} subset, with and without CASM. Fig. \ref{Abla_casm} clearly shows that CASM boosts the performance for each of the classes, especially for the minor classes, \eg, with an improvement of 2.16\% and 2.12\% AP on `truck' and `awning tricycle', respectively. This demonstrates the effectiveness of CASM, which uplifts the occurrence for minor classes during the training phase, resulting in more supervisory signals and consequently, achieving higher accuracy.

\noindent\textbf{Evaluation on other detection algorithms} We first instantiate the proposed adaptive feature selection module into the CenterNet detection framework, to evaluate its contribution. However, we emphasize that AFSM is a plug and play module, which can also be applied to other detection algorithms, irrespective of one-stage or two-stage. Therefore, more experiments are carried out on two representative detectors, Faster R-CNN \cite{ren2015faster} and CornerNet \cite{law2018cornernet}. The former is a state-of-the-art two-stage detector and, the latter is the first to attempt to detect objects with the keypoint estimation method.

The experimental results are showed in Table \ref{Abla_eva_others}. AFSM consistently increases the performance of two selected detection algorithms, with ResNet50 and ResNet101 as the backbone network. However, it is worth to note that the contribution to CornerNet is more significant, which improves from 72.13\% to 73.60\%, an increase of 1.47\% mAP, when applying AFSM to CornerNet. This is due to the fact that, CornerNet uses a single-scale feature (\ie, level 3) for the subsequent detection head, and AFSM facilitates the detection branch considerably, with the ability to adaptively selecting useful information from various feature scales. On the other hand, Faster R-CNN has already employed FPN for multi-scale object detection, while AFSM can still improve the performance by 0.43\% mAP, demonstrating the effectiveness of AFSM.

\begin{table}[tbp]
	\centering
	\renewcommand\tabcolsep{8pt} 
	\caption{Contribution of AFSM to CornerNet \cite{law2018cornernet} and Faster R-CNN \cite{ren2015faster}. mAP (\%) is reported on VOC 2007 test set. $(800, 1333)$ implies that the shorter size of image is resize to 800, while keeping the longer size within 1333. $\dagger$ means that ResNet50-D \cite{he2019bag} is used as backbone network.}
	\label{Abla_eva_others}
	\begin{tabular}{c|c|c|c|c|c}
		\toprule
		Method       & Backbone  & Resolution & AFSM & mAP   & FPS   \\ \midrule
		CornerNet    & ResNet50$\dagger$  & 511$\times$511    &      & 72.13 & 31.5  \\
		CornerNet    & ResNet50$\dagger$   & 511$\times$511    & \checkmark  & 73.60  & 30.77 \\
		CornerNet    & ResNet101 & 511$\times$511    &      & 70.67 & 29.2  \\ 
		CornerNet    & ResNet101 & 511$\times$511    & \checkmark  & 71.44 & 27.9  \\ \midrule[0.75pt]
		Faster R-CNN & ResNet50$\dagger$   & (800,1333) &      & 81.63 & 25.23 \\
		Faster R-CNN & ResNet50$\dagger$   & (800,1333) & \checkmark  & 82.06 & 21.86 \\ \bottomrule
	\end{tabular}
\end{table}

\section{Conclusion}
\label{conclusion}
This paper aims at tackling two fundamental challenges that limit the performance of object detection algorithms, \ie, the wide range of variation in object scale and the class imbalance problem. To address these challenges, a novel and effective adaptive feature selection module (AFSM) is first proposed, which adaptively learns the selection weight at each scale in the channel dimension, to enhance the important clues at a certain feature level. Secondly, we propose a class-aware sampling mechanism (CASM) to automatically assign the sampling weight to each of the images during training, according to the number of objects in each class. Each of the novel designs can improve the performance of the detector considerably. Finally, we uplift the baseline to a new state-of-the-art performance, with tiny inference overhead, thanks to the novel proposed methods.


\section*{Acknowledgment}



\bibliographystyle{elsarticle-num}

\bibliography{sample}

\end{document}